\documentclass[10pt,twocolumn,letterpaper]{article}

\usepackage[pagenumbers]{cvpr} 

\definecolor{cvprblue}{rgb}{0.21,0.49,0.74}
\usepackage[pagebackref,breaklinks,colorlinks,allcolors=cvprblue]{hyperref}

\def\paperID{*****} 
\def\confName{CVPR}
\def\confYear{2025}

\title{\methodname: Retrieval Augmented Video Generation for Enhanced Motion Realism}

\author{Elia Peruzzo\textsuperscript{1,*} \quad Dejia Xu\textsuperscript{2} \quad Xingqian Xu\textsuperscript{3,4} \quad Humphrey Shi\textsuperscript{3,4} \quad Nicu Sebe\textsuperscript{1}\\
{\small \textsuperscript{1}University of Trento \quad \textsuperscript{2}UT Austin \quad \textsuperscript{3}SHI Labs @ Georgia Tech \& UIUC \quad \textsuperscript{4}Picsart AI Research}}

\newcommand{\video}{\mathcal{X}}
\newcommand{\videonn}{\mathcal{Z}}
\newcommand{\prompt}{\mathcal{T}}
\newcommand{\videooutput}{\mathcal{Y}}

\newcommand{\dataset}{\mathcal{D}}
\newcommand{\query}{q}
\newcommand{\knumber}{K}
\newcommand{\qvideo}{\mathbf{x}}
\newcommand{\qprompt}{\mathbf{t}}

\newcommand{\ftext}{f_{\text{txt}}}
\newcommand{\fvis}{f_{\text{vis}}}
\newcommand{\findex}{f_{\text{index}}}
\newcommand{\vqenc}{\mathcal{E}}
\newcommand{\vqdec}{\mathcal{D}}
\newcommand{\unet}{\epsilon_\theta}
\newcommand{\textenc}{\tau_\theta}

\newcommand{\retrieval}{f_K}
\newcommand{\generator}{g_\theta}

\usepackage{pifont}
\usepackage{colortbl}
\usepackage{mathtools}
\usepackage{multirow}
\usepackage{graphicx}
\usepackage[export]{adjustbox}
\usepackage{xcolor}
\usepackage{xspace}

\newcommand{\methodname}{\textsc{RagMe}}
\newcommand{\supp}{\emph{Supp.~Mat.}}

\newcommand{\RNum}[1]{\uppercase\expandafter{\romannumeral #1\relax}}

\definecolor{myblue}{rgb}{0.458, 0.7333, 0.9921} 

\begin{document}
\maketitle

\begin{abstract}
Video generation is experiencing rapid growth, driven by advances in diffusion models and the development of better and larger datasets. However, producing high-quality videos remains challenging due to the high-dimensional data and the complexity of the task. Recent efforts have primarily focused on enhancing visual quality and addressing temporal inconsistencies, such as flickering. Despite progress in these areas, the generated videos often fall short in terms of motion complexity and physical plausibility, with many outputs either appearing static or exhibiting unrealistic motion. 
In this work, we propose a framework to improve the realism of motion in generated videos, exploring a complementary direction to much of the existing literature. Specifically, we advocate for the incorporation of a retrieval mechanism during the generation phase. The retrieved videos act as grounding signals, providing the model with demonstrations of how the objects move. Our pipeline is designed to apply to any text-to-video diffusion model, conditioning a pretrained model on the retrieved samples with minimal fine-tuning. We demonstrate the superiority of our approach through established metrics, recently proposed benchmarks, and qualitative results, and we highlight additional applications of the framework. 
\end{abstract}

\let\thefootnote\relax\footnotetext{*Corresponding author: {\texttt{elia.peruzzo@unitn.it}. Code available at: \url{https://github.com/helia95/ragme}.}}
\section{Introduction}
\label{sec:intro}
Text-to-video (T2V) generation is rapidly advancing, with large-scale models trained on vast datasets achieving increasingly impressive results. Notably, SORA \cite{videoworldsimulators2024} has established a new state-of-the-art, showcasing the remarkable potential of massive data and computational scaling. 
However, a significant limitation of current models lies in the realism and motion complexity of the objects in the output results. The generated videos often result in static scenes with simplistic or physically implausible motion \cite{opensora}. Some works tackle this issue by improving the data curation pipeline \cite{blattmann2023stable} or proposing a different architecture that scales better with the computation \cite{menapace2024snap}. However, all these models seem to suffer from similar failure cases, suggesting that scaling data and computing power are not sufficient to solve the problem.

\begin{figure}[t]
    \centering
    \includegraphics[width=\columnwidth,keepaspectratio]{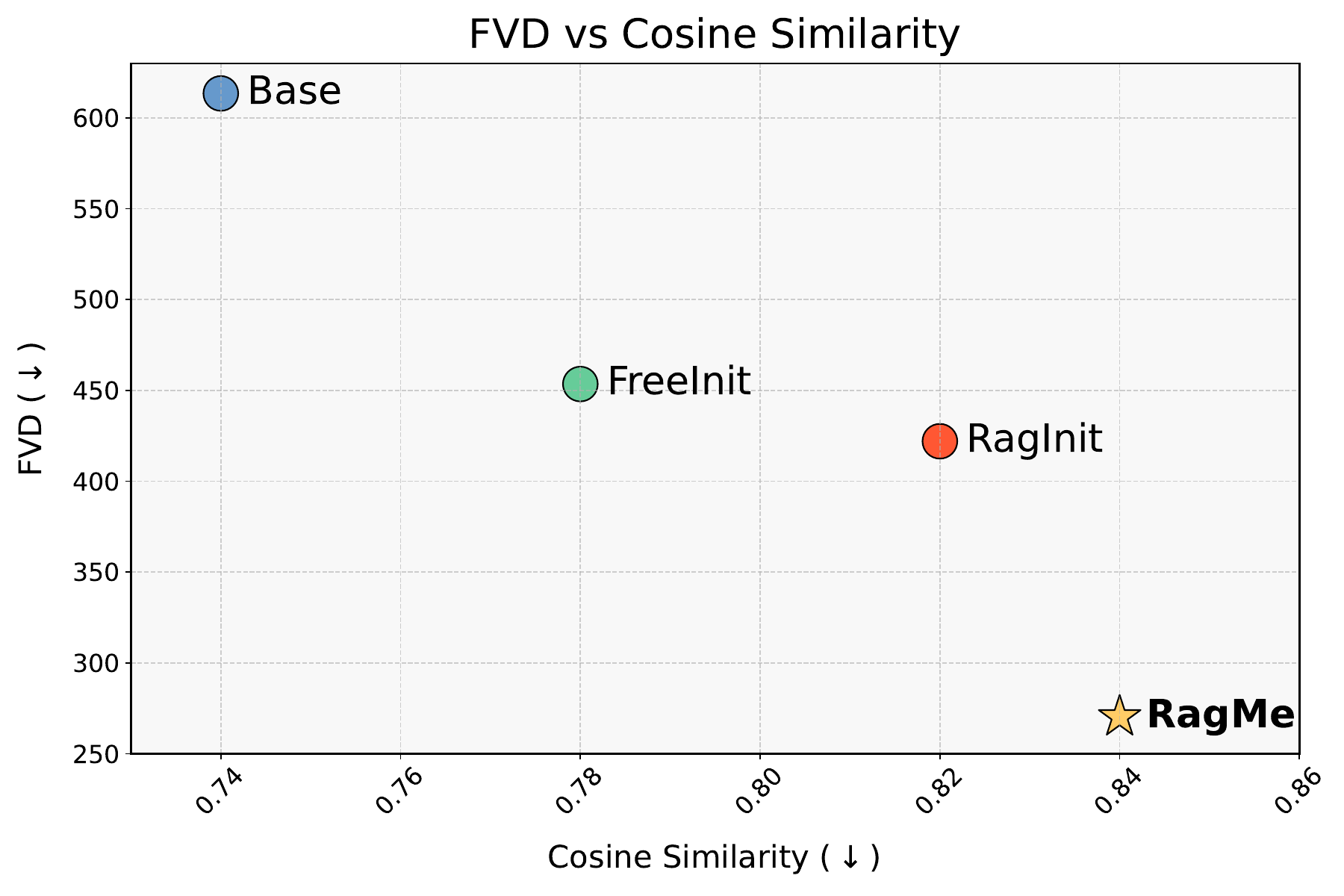}
    \caption{We evaluate the Fréchet Video Distance (FVD) using the captions and videos from the validation set of the WebVid10M \cite{Bain21} dataset. We plot it against the cosine similarity with respect to the retrieved examples in the DINOv2 embedding space. Ideally, the best model should produce high-quality videos (indicated by low FVD) while avoiding direct copying from the grounding examples (indicated by low cosine similarity).} 
    \label{fig:teaser}
\end{figure}

In this work, we explore a complementary approach, i.e.,  incorporating grounding information to guide the network toward a more realistic and plausible motion. We propose a retrieval augmented generation (RAG) pipeline -- a technique that has demonstrated impressive results in Natural Language Processing (NLP) \cite{lewis2020retrieval, 10.1162/tacl_a_00605}. However, it remains underutilized in computer vision, particularly in video generation. We retrieve (real) examples from an external database to guide the model and enhance the temporal dynamics of the generated samples. We term our method \methodname{}, Retrieval Augmented Generation for Motion Enhancement.

Our approach is inspired by the related tasks of video editing and motion transfer \cite{materzynska2023customizing, qi2023fatezero, yang2023rerender, esser2023structure}. In these settings, the goal is to synthesize an output video given one (or more) input video and a prompt describing the edit. The input videos are crucial for preserving motion, serving as an anchor for the video editing algorithm. We draw from these techniques but apply them to the broader problem of video generation. Our goal is to transfer the high-level action from the retrieved examples without preserving their specific details. Specifically, our design choices focus on preventing the transfer of low-level details, such as the background, the subject's identity, or the spatial arrangement of the scene.
For example, when generating a video of a person walking, we can gather samples from an external database where the action is performed in various ways. People have distinct identities and walk in different ways, in different directions, and across different environments. 
However, the underlying action remains consistent across these examples, and all of these variations can guide the model to produce a video with a more realistic motion.
In this work, we aim to preserve only high-level information, allowing the model to generate new content without directly copying specific instances from the retrieved examples. When evaluating Fréchet Video Distance (FVD), our method significantly reduces this metric compared to the base model while ensuring that the generated video is not a replica of the retrieved samples, as indicated by a slight increase in cosine similarity between them (see \Cref{fig:teaser}).

We build our pipeline in a general manner, without specific assumptions about the architecture or the application (\eg, humans). We use the WebVid10M as a large-scale text-to-video dataset and use it to build a retrieval mechanism, which is used to condition a pre-trained T2V model by inserting cross-attention layers that fuse information from retrieved samples. Additionally, we propose a novel mechanism to initialize the random noise for the denoising process leveraging the retrieved samples. We evaluate our model through standard metrics like FVD, but also on the recently proposed video generation benchmarks. We demonstrate superior results compared to baselines and training-free methods for enhancing video quality and consistency.
The core contribution of this work is to apply for the first time a RAG pipeline to video generation as a first step to guide the model towards more realistic motion generation.

\section{Related Works}
\paragraph{Text-to-Video Diffusion Models}
In the last years, there have been several efforts to expand the achievements of text-to-image models to the video domain \cite{singer2022make, ge2023preserve, ho2022imagen, Blattmann_2023_CVPR, wang2023modelscope, wang2023lavie}. ImagenVideo \cite{ho2022imagen} and Make-A-Video \cite{singer2022make} propose a deep cascade of temporal and spatial upsamplers to generate videos and jointly train their models on image and video datasets. 
A consistent line of works focus on extending powerful pre-trained text-to-image (T2I) models introducing new layers to model the time dimension and exploiting the powerful prior learned on the spatial domain \cite{wang2023lavie, wang2023modelscope}. 
Blattmann \etal \cite{blattmann2023align} initially explored this direction by leveraging a pre-trained Stable Diffusion model \cite{rombach2022high}, which was later extended to image-to-video generation and longer videos by Stable Video Diffusion \cite{blattmann2023stable}.  AnimateDiff \cite{guo2023animatediff} proposes to freeze the spatial layers and train only the temporal module and introduce MotionLoRA \cite{hu2021lora} as a lightweight finetuning technique to learn specific motion patterns. Nevertheless, all these methods rely on 3DUNet with separable spatial and temporal computation which poses a limitation on motion modeling capabilities. SnapVideo \cite{menapace2024snap} proposes to use a transformer-based FIT \cite{li2023fit} architecture which can jointly model the space and time components, by exploiting a compressed video latent representation. Other works introduce fully transformer-based architectures \cite{ma2024latte}, culminating in the state-of-the-art results achieved by SORA \cite{videoworldsimulators2024}. While the open-source community is working to replicate these outcomes, the generated quality still lags behind \cite{opensora, pku_yuan_lab_and_tuzhan_ai_etc_2024_10948109}.

Concurrently, some approaches have explored not only the architectural modeling choices but also the noising policy. Pyoco \cite{ge2023preserve} introduces a noise-correlated sampling strategy, based on the intuition that frames shouldn't be sampled from independent noise. Recently, FreeInit \cite{wu2023freeinit} proposed a training-free technique to optimize the initial noise of the denoising process. The model predicts a sample that is diffused back according to the noising schedule, mixing the low-frequency components with randomly initialized high-frequency components. While this approach results in improved sample consistency, it requires repeating the sampling process multiple times, which is often impractical.

\emph{We build on the recent advancement of T2V models, leveraging the strengths of powerful pre-trained models and extending their capabilities with minimal architecture modifications. Additionally, we propose a noise initialization strategy that enhances the final result without incurring the high computational costs associated with existing methods}.

\paragraph{Motion Transfer and Video Editing}
One line of work exploits pre-trained T2I models and adapts them to the task in a zero-shot manner \cite{qi2023fatezero, ceylan2023pix2video, khachatryan2023text2video, geyer2023tokenflow, yang2023rerender}. The temporal consistency of the generated frames is typically obtained by extending the self-attention operation across frames\cite{khachatryan2023text2video, wu2023tune}. Tune-A-Video \cite{wu2023tune} involves fine-tuning the model on the video to be edited, enabling test-time edits through text prompts or cross-attention control \cite{liu2023video}. Pix2Video \cite{ceylan2023pix2video} and FateZero \cite{qi2023fatezero}  propose a training-free approach, exploiting the attention maps extracted during an initial inversion step and blended with those generated during the editing process, confining the edit to a specific region. TokenFlow \cite{geyer2023tokenflow} and FLATTEN \cite{cong2023flatten} propose to propagate features of the base T2I model leveraging the optical flow extracted from the source video. In contrast, other methods opt for pretraining on video datasets, typically employing an inflated 3DUNet architecture and incorporating explicit dense conditioning signals (e.g., optical flow, depth maps, or sketches) to preserve motion and structure from the guiding video \cite{esser2023structure, wang2023videocomposer, guo2023animatediff, guo2023sparsectrl, peruzzo2024vase}. Animate-A-Story \cite{he2023animate} utilizes a similar technique for guiding generation, but instead of relying on user-provided input, it retrieves a single video from a database to serve as the anchor.
Other works have explored the broader task of motion transfer. Yatim \etal \cite{Yatim_2024_CVPR} addresses motion transfer between objects of different categories that may not share the same motion characteristics. They enforce the transfer through an inference-time optimization, introducing a loss to match the correlation of features of the input with the output video. Similarly, \cite{materzynska2023customizing, zhao2024motiondirector} propose a DreamBooth-like \cite{ruiz2023dreambooth} training strategy to learn motion patterns from a set of videos with the same action.

\emph{Our work is inspired by this line of research but differs fundamentally because we do not aim to replicate the conditioning video, nor do we rely on a manually curated set of examples. Furthermore, we seek a practical implementation that avoids costly test-time training procedures.}

\paragraph{Retrieval Augmented Generation (RAG)} It represents a well established technique in Natural Language Processing as a powerful way to improve model performances, by integrating information from an external database that acts as a memory bank \cite{lewis2020retrieval, 10.1162/tacl_a_00605, pmlr-v162-borgeaud22a}. 
Early attempts to adapt similar retrieval mechanisms for image and video generation were introduced within the context of GANs \cite{tseng2020retrievegan, casanova2021instance}. More recently, \cite{blattmann2022retrieval, sheynin2022knn} have applied these concepts to image diffusion models. Their approach involves a semi-parametric generative model that combines a learnable module with an external database, allowing for post-hoc conditioning based on labels, prompts, or specific styles. Re-Imagen \cite{pmlr-v162-borgeaud22a} extends this concept to text-to-image (T2I) models, and \cite{yang2024mastering} propose an in-context learning strategy to integrate retrieved samples and enhance generation results.

\emph{To the best of our knowledge, RAG has not yet been applied to video generation, which presents additional challenges in both the retrieval mechanism and the model's conditioning component.}

\section{Method}
\label{sec:method}
We describe the technical details of \methodname{}, formalize the task, and outline its applications. 
We begin by defining the notation used throughout the paper.
We assume to have access to a database $\mathcal{D} = \{\video_i\}_{i=1}^N$. Each data-point represents a video, with $\video_i \in \mathbb{R}^{T \times 3 \times H \times W}$ denotes the $T$ frames of the video with spatial resolution $H \times W$.

We define a \emph{Retrieval Mechanism} (RM) as a non-learnable function to retrieve from the database given a query $\query$, \ie $\retrieval: (\query, \mathcal{D}) \rightarrow \mathbf{Z}$, with $\mathbf{Z} = \{(\video_j, \prompt_j)\}_{j=1}^\knumber$, $\mathbf{Z} \subseteq \dataset$ and $\knumber = \lvert \mathbf{Z} \rvert$ represents the number of retrieved samples. 
Next, we define $\generator: \prompt_i \rightarrow \videooutput_i$ as a (pretrained) \emph{T2V Generative Model} that synthesizes an output video $\videooutput_i \in \mathbb{R}^{T \times 3 \times H \times W}$ given a textual prompt $\prompt_i$.

In this work, we propose to learn a \emph{semi-parametric} T2V model, which can incorporate relevant retrieved samples via conditioning, \ie $g_{\theta'} : (\prompt_i, \mathbf{Z}) \rightarrow \videooutput_i$. As discussed in \cref{sec:intro}, our final goal is to produce videos with better temporal dynamics, \emph{without copy-pasting} artifacts from the retrieved examples. 


\begin{figure*}[t]
    \centering
    \includegraphics[width=\linewidth]{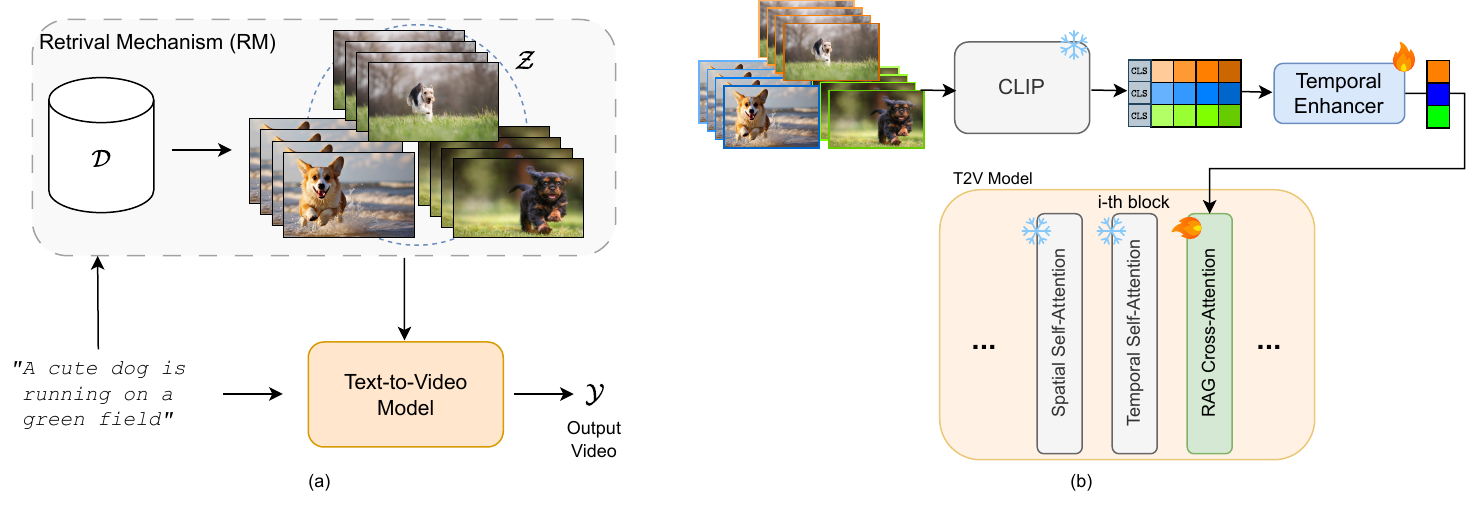}
        \caption{Pipeline of \methodname. (a) We show a general T2V pipeline with RAG capabilities. Given a textual prompt, we retrieve related videos from a database and use it to enhance the generation capabilities of a T2V model. (b) We detail the specific implementation. Each video frame from the retrieved videos is encoded using CLIP and then processed by a transformer temporal enhancer module to obtain the final conditioning vector. This vector is used to condition a T2V model through cross-attention layers. Each video is color-coded, with different frames represented by varying shades of the base color.}
    \label{fig:method}
\end{figure*}

\paragraph{T2V Diffusion Models Preliminaries} 
Diffusion models are probabilistic models that approximate distributions by iteratively denoising data. Starting with a sample of Gaussian noise, the model learns to progressively remove noise in steps until the sample approximates the target distribution \cite{ho2020denoising, song2020denoising}. Our framework builds upon a pre-trained \emph{latent} T2V model \cite{rombach2022high, blattmann2023stable}. Instead of learning the distribution directly in the complex, high-dimensional video space, this model projects the video into a compressed latent representation and learns a conditional distribution based on text.
Architecturally, it consists of three main components: The VAE Encoder $\vqenc(\cdot)$, which projects the raw input pixels to the latent space \ie $z = \vqenc(\video)$, and the correspondent Decoder $\vqdec(\cdot)$. The text encoder $\textenc(\cdot)$, which maps the input textual prompt to a conditioning vector; and the denoiser $\unet(\cdot)$, which takes the text embedding and a noisy version of the latent as input and predicts (with the correct reparametrization \cite{ho2020denoising}) the added noise.

The training is performed by sampling a noise $\epsilon \sim \mathcal{N}(0, 1)$ and diffusing the original sample $z_0$ according to a noise scheduler function and a time-step $t \sim f(t)$ \cite{ho2020denoising, karras2022elucidating, dhariwal2021diffusion}. The diffused sample $z_t$ is computed as 

\begin{equation}
    z_t = \sqrt{\alpha_t} \cdot z_0 + \sqrt{1-\alpha_t} \cdot \epsilon
    \label{eq:diffuse}
\end{equation} 

where $\alpha_t$ is a parameter controlled by the noise scheduler function that dictates the amount of noise at timestep $t$. At the final timestep $t = T$, the original sample is completely destroyed to pure noise, \ie $z_T \sim \mathcal{N}(0, 1)$, which allows sampling from the model at inference time.

The parameters of the denoiser network are trained to recover the added noise. Specifically, the training loss is defined as:

\begin{equation}
    \label{eq:train_loss}
    \mathcal{L}_{\text{simple}} \coloneqq \mathbb{E}_{\vqenc(x), \epsilon\sim\mathcal{N}(0,1),t}\Big[\lVert \epsilon - \unet(z_t, t, \textenc(c) \rVert_2^2 \Big]
\end{equation}

In this work, we focus on the denoiser network $\epsilon_\theta(\cdot)$. Although purely transformer based architecture are emerging, we rely on the widespread 3DUNet models \cite{blattmann2023stable, wang2023modelscope, Blattmann_2023_CVPR, wang2023lavie}. From an architectural perspective, combines convolutional layers with attention operations. The attention blocks can be further categorized into the:

\begin{itemize}
    \item \emph{Cross-Attention} blocks, which integrate information from the text encoder.
    \item \emph{Spatial Attention} blocks, which operate on the spatial dimension treating each frame independently, the activation of the network are reshaped as $x_{\text{spatial}} \in \mathbb{R}^{(b \cdot T) \times (h \cdot w) \times \text{dim}}$.
    \item \emph{Temporal Attention} blocks, which operate solely on the temporal axis, the activation of the network are reshaped as $x_{\text{temp}} \in \mathbb{R}^{(b \cdot h \cdot w) \times T \times \text{dim}}$.
\end{itemize}

In this work, we concentrate on the \emph{temporal attention} blocks, as our primary goal is to enhance the temporal dynamics of the generated video.

\paragraph{Retrieval Mechanism (RM)} The retrieval mechanism processes a query $\query$ and retrieve $\knumber$ samples form a database $\mathcal{D}$. The retrieval is performed by minimizing a distance function $d(\query, \cdot)$ between the query and the other entries in the database. In practice, it is composed of three non-learnable blocks: the pre-trained text encoder $\ftext$, the pre-trained visual encoder $\fvis$ and an indexing mechanism $\findex$. Following the previous works, we use CLIP to implement the visual and textual encoders. Our choice is motivated by three factors: (i) previous works on video-action recognition show that frame-wise CLIP encodings are powerful for the task, and can be used to recognize the action with high accuracy \cite{Bain21, wang2021actionclip, luo2022clip4clip, ma2022x} (ii) the embedding space is compact and reduces the dimensionality ($\text{dim} = 512$), with advantages in memory and computational requirements, (iii) the shared textual-visual embedding space allows to search the database in a multi-model manner at inference time (\ie using the prompt of the T2V model as the query for the retrival) \cite{blattmann2022retrieval}. %

First, we preprocess the database $\dataset$. For each video $\video_i$, we encode the frames independently and compute the average along the temporal dimension to aggregate the information. This results in a per-video representation, after L2 normalization:

\begin{equation}
    \label{eq:video}
    \qvideo_i  = \Big\lVert\frac{1}{T} \sum_{j=1}^T \lVert\fvis(\video_{i,j})\rVert_2\Big\rVert_2. 
\end{equation}

Second, we efficiently store the compressed video representations in the index using the FAISS library \cite{douze2024faiss}. Next, we search over the index, returning $\knumber$ samples from the database, which maximize the \emph{cosine similairty} $d_{\text{cos}}$ with the query:

\begin{equation}
    \label{eq:search}
    \mathbf{Z} = \underset{\videonn_j \in \dataset}{\texttt{top-\knumber}} \quad d_{\text{cos}}(\query, \videonn_j)
\end{equation}
with $\mathbf{Z} = \{\videonn_0, \ldots, \videonn_K\}$, $q \notin \mathbf{Z}$.

During training, we compute the averaged temporal CLIP representation for the current video $\video_i$ as described in \cref{eq:video}. Then, we search the dataset using \cref{eq:search}, setting the query $\query = \qvideo_i$. Conversely, at test time, we encode the given textual prompt $\prompt_i$ using the CLIP textual encoder, \ie, $\qprompt_i=\lVert\ftext(\prompt_i)\rVert_2$. Finally, we leverage the multimodal nature of the CLIP latent space and retrieve from the dataset using \cref{eq:search}, setting the query $\query=\qprompt_i$. We refer to \cref{fig:method} (a) for a visual representation of the process.

Note that, for the sake of generality, we assume the database to contain \emph{only} videos, but the pipeline can be applied to text-video database as well. We explore other choices for the retrieval system and discuss the result in the \Cref{sec:experiments}. Lastly, 
we apply a deduplication strategy to prevent returning (multiple) similar elements in a dataset with redundant entries. Further details on the implementation and post-processing are provided in the \supp.

\paragraph{Retrieval Augmented Conditioning (RagCA)}
After developing the retrieval mechanism, we explain how to condition the T2V model using this retrieved information. For a visual representation of the process, refer to \cref{fig:method} (b). The first step involves representing the conditioning videos within an appropriate embedding space. Consistent with our guiding principle, our goal is to condition the main network in a way that enhances temporal dynamics, while avoiding direct copies of the the conditioning signals. The CLIP visual encoder emerges as a strong candidate for this purpose, as it effectively encodes high-level semantic without retaining low-level information \cite{radford2021learning}. Additionally, it offers a practical solution since we can directly utilize the embeddings returned by the retrieval mechanism. However, since $\fvis$ operates on independent frames, we introduce a module specifically designed to handle the temporal dimension, which we term the \emph{transformer time enhancer} model. In practice, we pack the per-frame CLIP embedding into a sequence of tokens:

\begin{equation}
    \label{eq:transformer}
    \bar{\mathbf{z}}_i = [\texttt{CLS}; \fvis(\videonn_{i,0}); \ldots; \fvis(\videonn_{i, T})]  
\end{equation}

with $\bar{\mathbf{z}}_i\in \mathbb{R}^{(T+1) \times \text{dim}}$, 
$[\ldots;\ldots]$ represents the concatenation operation and $[\texttt{CLS}]$ is a class token appended at the beginning of the sequence \cite{dosovitskiy2020vit}. We apply the transformer time enhancer independently on each retrieved videos and pool the $[\texttt{CLS}]$ token in output. In this way, we obtain the final conditioning signal $\mathbf{z} = \tau(\bar{\mathbf{z}})$, with $\mathbf{z} \in \mathbb{R}^{b\times K \times\text{dim}}$ (see \cref{fig:method} (b)). 

Next, we condition the pre-trained T2V model retaining the generation capabilities learned during the pertaining stage. Following previous works, we initialize new multi-head cross attention layers and inject them after every temporal attention layer of the base model. In practice, let $x_{\text{temp}} \in \mathbb{R}^{(b \cdot h \cdot w) \times T \times \text{ch}}$ be the 3DUNet activation after a temporal layer, we compute a residual:

\begin{equation}
    x_{\text{temp}} = x_{\text{temp}} + \texttt{MCA}(x_{\text{temp}}, \mathbf{z})
    \label{eq:conditioning}
\end{equation}

where $\texttt{MCA}(\cdot)$ represent the multi-head cross-attention operation with queries computed from $x_{\text{temp}}$ and keys/values from the $\mathbf{z}$ signals respectively. 

\paragraph{RAG Noise Initialization (RagInit)} As explored in previous works \cite{wu2023freeinit, karthik2023if, xu2024good}, noise initialization plays an important role in diffusion models and can greatly affect the quality of the generated result. We further leverage the retrieved videos and propose to initialize the noise averaging the latents. We diffuse the result following \cref{eq:diffuse} and setting $t=T$:

\begin{equation}
        z_T^{\text{RAG}} = \sqrt{\alpha_T} \cdot \dfrac{1}{K} \sum_{i=1}^K\vqenc(\videonn_i) + \sqrt{1-\alpha_T} \cdot \epsilon
\end{equation}

This strategy is very fast, as it doesn't require inversion, and comes at the additional cost of running the VAE encoder on the retrieved videos. Nevertheless, it has the advantage of providing a good initialization for the noise which is likely to be aligned with the conditioning videos. 


\paragraph{Implementation Details} 
We build our framework on Zeroscope \cite{zeroscope}, a latent T2V model based on an inflated 3DUNet architecture with factorized spatial and temporal layers. We develop the retrieval system using the WebVid10M dataset \cite{Bain21}; our choice is motivated by the large scale and the general-purpose nature of its videos, which cover a wide range of scenarios. For the retrieval mechanism, we implement the CLIP ViT-B-32 \cite{radford2021learning} as our feature extractor to handle both $\fvis$ and $\ftext$. This model, pre-trained with a contrastive loss on images and captions from a large-scale dataset, outputs a 512-dimensional embedding representing the respective input. Although the choice of the encoder for the retrieval mechanism could, in principle, be independent of the conditioning process, we find it easier and more convenient to use the same encoder.

Next, we leverage the FAISS library \cite{douze2024faiss} to create an index for efficient retrieval. The WebVid10M dataset contains duplicate or highly similar videos; to prevent the model from processing redundant information, given a query $\mathbf{q}$, we apply a deduplication strategy based on the cosine similarity between samples. We empirically set the deduplication threshold at $\delta_{\text{dedup}} = 0.965$ and maintain this value across all experiments. Additionally, to ensure that the retrieved videos are relevant to the query, we set a minimum cosine similarity threshold of $\delta_{\min} = 0.6$ and remove samples from the retrieval set $\mathbf{Z}$ that do not meet this criterion. This filtering is particularly applied when retrieving a large number of samples (\ie $K = 20$, $K = 50$). 
In such cases, padding is used to match the required length.

From an architectural point of view, we introduce the transformer temporal enhancer module to improve the temporal representation of the video. It is composed of 6 layers of transformer blocks with a hidden dimension of $\text{dim} = 512$. A learnable token $\texttt{[CLS]}$ is added at the beginning of the sequence and pooled in output to represent the video. Lastly, we add multi-head cross-attention layers to the base T2V model ZeroScope. We introduce a point-wise convolution initialized with zero-weights, to act as the identity when the model is initialized. 

The added modules are finetuned (while keeping the rest of the network frozen) for 200K iterations on the WebVid10M dataset, at resolution $448 \times 256$ and $12$ frames. Training is performed with an effective batch size of 16, distributed on 4 Nvidia A100 GPUs.

\section{Experiments}
\label{sec:experiments}
In this section, we qualitatively and quantitatively analyze the performance of \methodname. We start by evaluating established metrics in the video generation field on the validation set of WebVid10M \cite{Bain21}. Moreover, we follow VBench \cite{huang2023vbench}, a benchmark recently introduced, which exploits an array of pre-trained models to evaluate the generated videos under multiple angles. Next, we present a series of ablation studies to understand the role of each component in our pipeline. Lastly, we showcase several qualitative results comparing our method with the baselines.

\paragraph{Baselines and Setting} We compare \methodname{} with videos produced by the base T2V model, ZeroScope \cite{zeroscope}. Next, we enhance the videos generated by the base model using FreeInit \cite{wu2023freeinit}, a training-free technique that optimizes the starting noise of the diffusion process through repeated denoising. Finally, we compare our full model with another baseline, which uses our proposed RagInit technique to initialize the noise.

We perform inference from all the models using the DDIM sampler \cite{song2020denoising} with 50 denoising steps, and classifier-free guidance with scale of $s = 7.5$.

\subsection{Quantitative Results}
\begin{table}[t]
    \centering
    \resizebox{\columnwidth}{!}{
    \begin{tabular}{lccc}
    \toprule
    \textbf{Method} &  FVD $(\downarrow)$ & DINO-S $(\downarrow)$   & Latency (s) $(\downarrow)$\\
    \midrule
    \rowcolor{gray!20} Retrieved Videos & 117.22 & 1.00 & - \\
    ZeroScope &  613.15  & 0.74   & 17.78 \\
    FreeInit  & 453.50  &  0.79 &  68.88 \\
    RAGInit &   422.10  & 0.82 &  19.86 \\
    \rowcolor{myblue!20} \textbf{\methodname{}} &  270.26 & 0.84 & 22.43 \\
    \bottomrule
    \end{tabular}}
    \caption{Comparison between the baseline methods and \methodname{} on the WebVid10M validation set.}
    \label{tab:webvid_validation}
\end{table}

\begin{table*}[thb]
    \centering
    \resizebox{\textwidth}{!}{
    \begin{tabular}{lccccccc}
    \toprule
    Method &  Human Action &  Subject Consistency &  Background Consistency &  Motion Smoothness &  Temporal Flickering &  Dynamic Degree \\
    \midrule

ZeroScope & 0.922 & 0.962 & 0.984 & 0.985 & 0.986 & 0.367 \\
FreeInit & 0.912 & \textbf{0.978} & \textbf{0.990} & \textbf{0.988} & \textbf{0.994} & 0.242 \\
RagInit (Our) & 0.952 & 0.961 & 0.985 & 0.985 & 0.990 & 0.467 \\
\rowcolor{myblue!20} \textbf{\methodname} & \textbf{0.974} & 0.911 & 0.972 & 0.968 & 0.982 & \textbf{0.692} \\

    \bottomrule
    \end{tabular}}
    \caption{Comparison between \methodname{}  and the baselines on VBench \cite{huang2023vbench}. We report the metrics related to motion dynamics and temporal consistency. Our method outperforms the competitors in the quality of motion while slightly decreasing the consistency-related metrics.}
    \label{tab:vbench}
\end{table*}

\paragraph{WebVid10M Results} Our end goal is to develop a system with better video quality, especially in the temporal dynamics, while avoiding leakage of the conditioning videos (see \cref{sec:intro}). To capture the first aspect, we rely on the Fr\'echet Video Distance (FVD) \cite{unterthiner2018towards}, which is well-established in the video generation literature. 
To estimate the second factor, \ie possible copy-paste artifact from the retrieved videos, we compute the cosine similarity on the DINOv2 \cite{oquab2023dinov2} embedding space. Specifically, given a generated video $\videooutput$ and a set of retrieved videos $\mathbf{Z}$, the metric is computed as $\max_\mathbf{Z} \text{cos-sim}(\videooutput, \videonn_i)$. In this case, a model that achieves a lower cosine similarity is considered better. Lastly, we compare the methods on the latency, \ie the time to generate a single video. We take into account the time of retrieving the videos and encoding them with CLIP when computing the latency of our model. We refer to the \supp{} for more detailed discussion.


We conduct the experiments on the WebVid10M validation set, which comprises 5000 videos with the associated captions. We report the results in \cref{tab:webvid_validation}, wherein the first row we report the results of the \emph{retrieved videos} (\ie videos form the WebVid10M training set) as a reference. \methodname{} drastically outperforms the base diffusion model in terms of FVD, resulting in videos of higher quality. While applying FreeInit does lead to some improvement,  it remains inferior in comparison. RagInit achieves comparable performance to FreeInit. However, a notable difference emerges in latency: our proposed noise initialization method does not require costly denoising steps and instead uses the retrieved samples for noise initialization.

Analyzing the DINO-similarity metric, we observe that \methodname{} shows an increase compared to both the baseline and FreeInit. However, compared to RagInit, the full model's improvement is minimal, suggesting that the primary issue may lie in the noise initialization procedure rather than the cross-attention conditioning. It is important to note that a \emph{very low} DINO cosine similarity is not desirable as well, and would indicate: either a lack of relevance between the retrieved videos and the final video or a failure of the T2V model to align with the prompt.

\paragraph{VBench Results} While the FVD metric is well established, it is difficult to interpret as improvements over it can be due to multiple factors. To get a better understanding of \emph{what aspects} our method is improving, we follow VBench \cite{huang2023vbench} for a more detailed evaluation. VBench is a recently proposed benchmark for T2V models, which comprises a suite of roughly 900 prompts and a list of 16 dimensions for evaluations. In the main paper, we report only the metrics related to the temporal consistency and quality of motion, as these represent our main target for improvement. However, we refer the reader to the \supp~for full comparison between the methods, and to the original paper for detailed explanation of how each metric is computed. We report the results in \cref{tab:vbench}. Our method strongly outperforms the baseline in two aspects: the Human Action and the Dynamic Degree metrics. This reflects our design goals of having less static videos with better motion. At the same time comes at the price of a slight decrease in background and subject consistency, which is nevertheless expected (a static video would achieve a perfect score in these metrics). Comparing with the noise initialization stargeies of FreeInit and RagInit, it is interesting to notice that a better action can be primarily explained by a better noise initialization, but the dynamic degree is mostly due to the corss-attention layers which incorporates the retrieved videos. 

\subsection{Ablations}

\begin{figure}
    \centering
    \includegraphics[width=\linewidth, keepaspectratio]{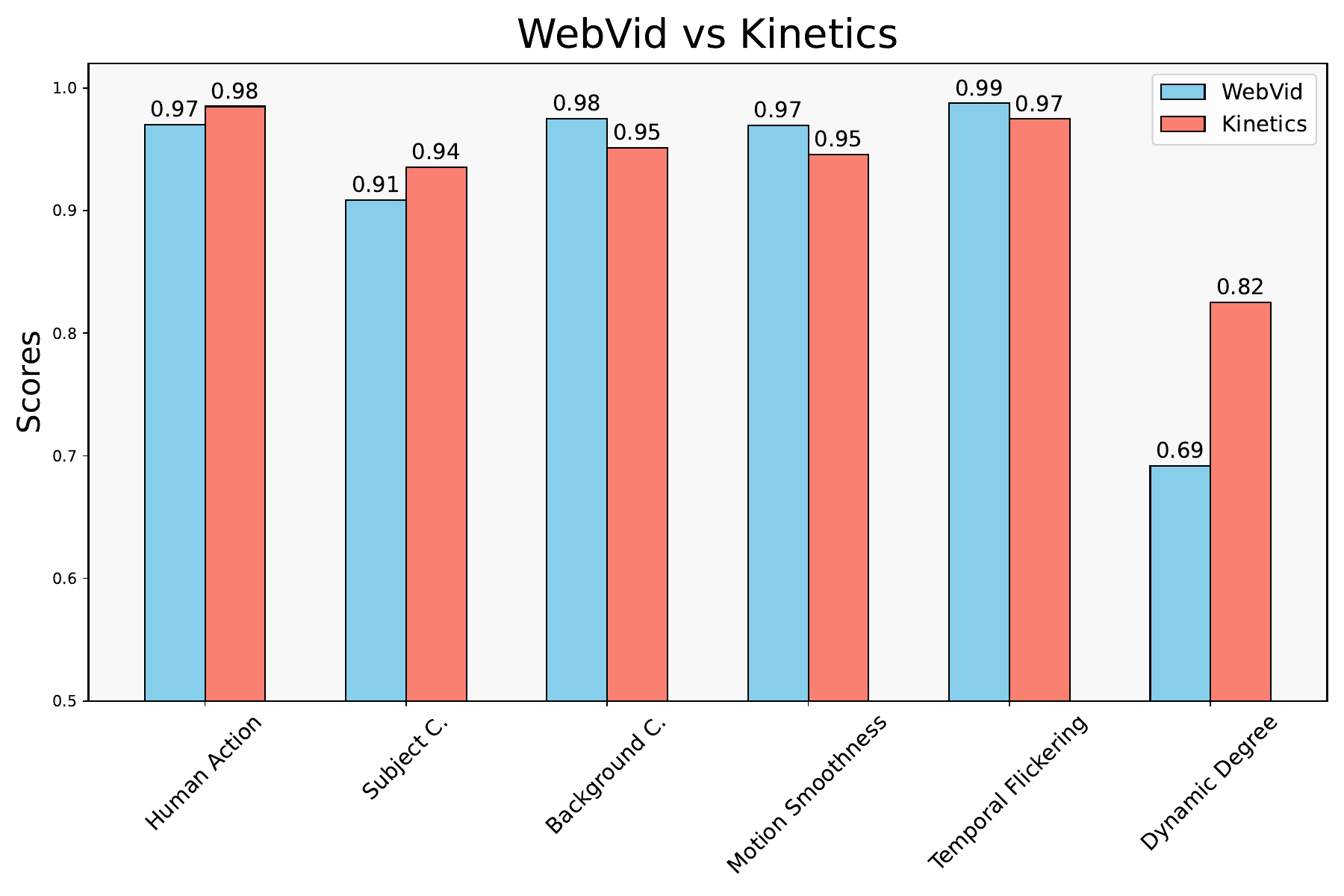}
    \caption{We compare the role of different retrieval databases on the person-related subset of VBench \cite{huang2023vbench}. We retrieve it from the Kinetics \cite{kay2017kinetics} and the WebVid10M \cite{Bain21}.} 
    \label{fig:kinetics_webvid}
\end{figure}

\paragraph{Role of the database $\mathcal{D}$}
We ablate the role of the retrieval database $\mathcal{D}$ in our system, specifically focusing on the types of videos we retrieve. In the previous section, we used a general retrieval mechanism without making strong assumptions about the task. The retrieval database consisted of general videos from WebVid, and we did not exploit the textual components. However, the proposed mechanism is highly flexible, allowing different databases to be used at inference time to retrieve videos tailored to specific applications. Hence, we assume access to an application-specific database for human-related prompts, specifically the Kinetics \cite{kay2017kinetics} video dataset, and plug it into our pipeline without further fine-tuning. This dataset, commonly used for action recognition tasks, contains a large set of actions performed by people. We replace our base dataset, derived from WebVid10M, with Kinetics and evaluate how this change affects performance on the VBench metrics. The results, shown in \cref{fig:kinetics_webvid}, demonstrate a relative improvement in both the Human Action and Dynamic Degree metrics. These findings highlight the importance of the retrieved videos in the process and suggest that the mechanism can be specialized for specific applications to achieve better performance.


\begin{figure}[!t]
    \centering
    \includegraphics[width=\linewidth]{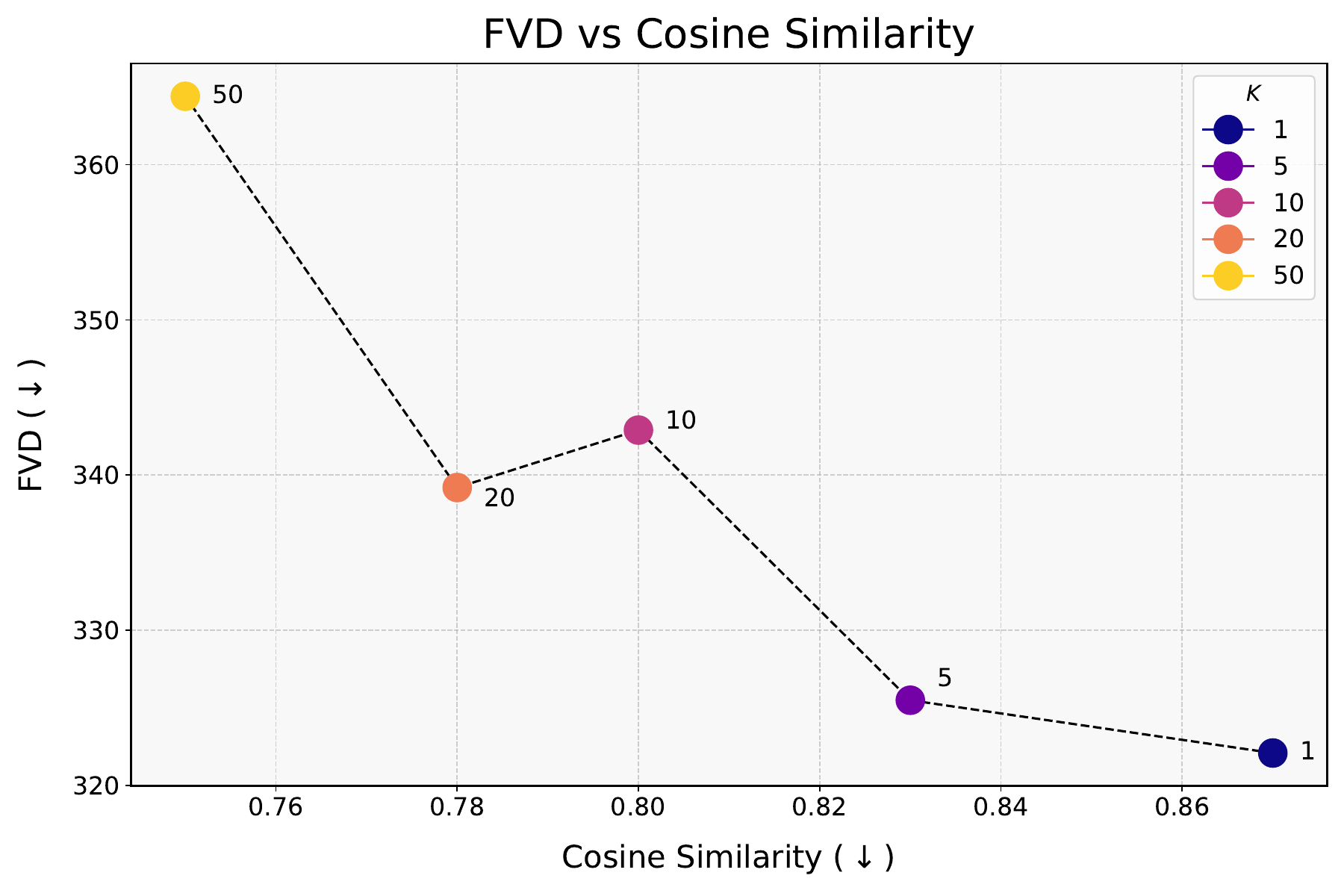}
    \caption{
    We study the impact of the retrieved samples $K$ on the FVD vs Cosine Similarity trade-off. We select $K=5$ as a good trade-off between the two.}
    \label{fig:role_of_k}
\end{figure}

\begin{table*}[!th]
    \centering
    \resizebox{\textwidth}{!}{
    \setlength\tabcolsep{0.3pt}
    \footnotesize
    \renewcommand{\arraystretch}{0.3}
    \begin{tabular}{cccc@{\hskip 1pt}ccc}
            \rotatebox{90}{\hspace{0.5mm}{\scalebox{.5}{ZeroScope}}} & 
            \includegraphics[width=0.15\columnwidth, frame]{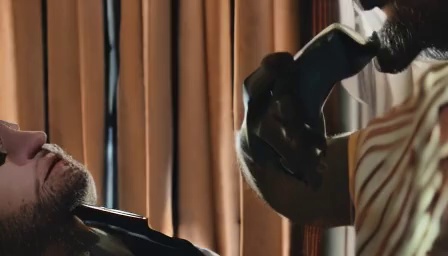} &
            \includegraphics[width=0.15\columnwidth, frame]{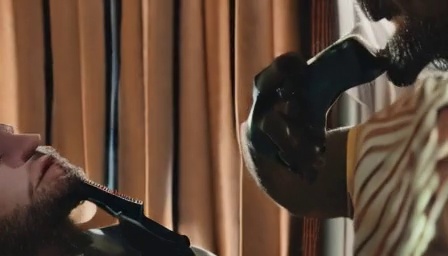} &
            \includegraphics[width=0.15\columnwidth, frame]{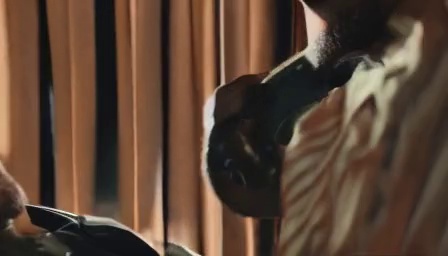} &
            \includegraphics[width=0.15\columnwidth, frame]{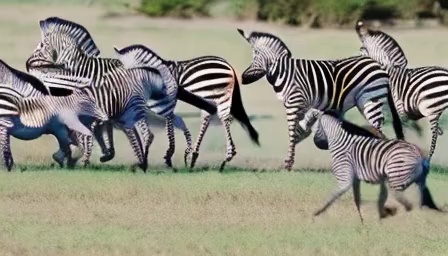} &
            \includegraphics[width=0.15\columnwidth, frame]{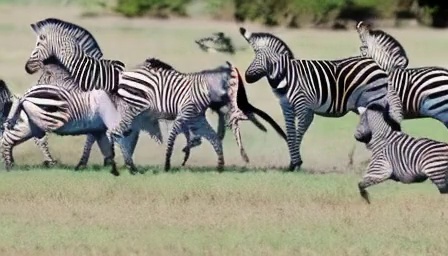} &
            \includegraphics[width=0.15\columnwidth, frame]{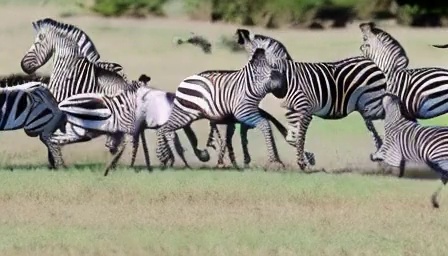}  \\

            \rotatebox{90}{\hspace{0.8mm}{\scalebox{.5}{FreeInit}}} & 
            \includegraphics[width=0.15\columnwidth, frame]{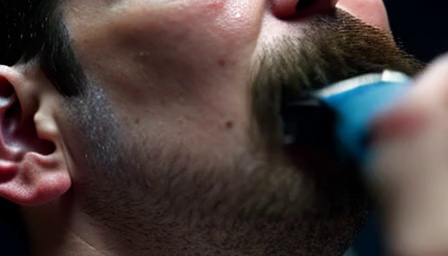} &
            \includegraphics[width=0.15\columnwidth, frame]{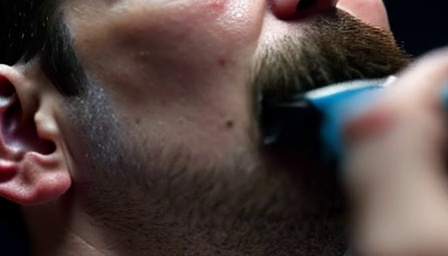} &
            \includegraphics[width=0.15\columnwidth, frame]{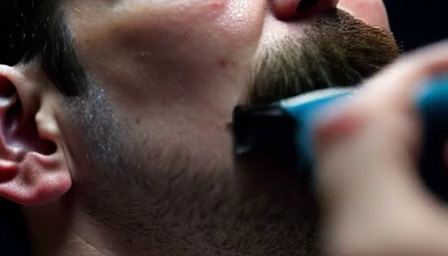} &
            \includegraphics[width=0.15\columnwidth, frame]{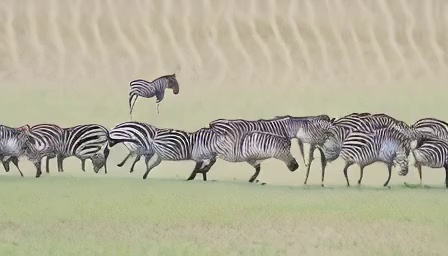} &
            \includegraphics[width=0.15\columnwidth, frame]{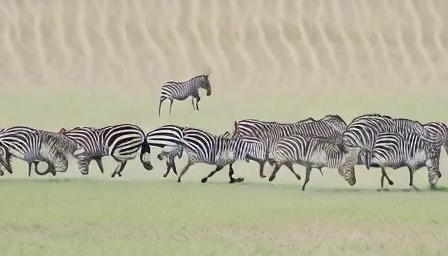} &
            \includegraphics[width=0.15\columnwidth, frame]{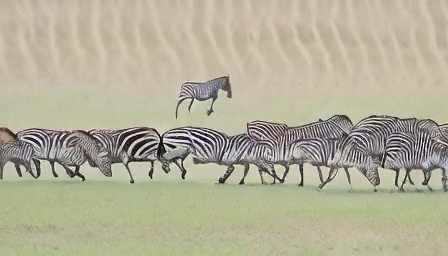}  \\

            \rotatebox{90}{\hspace{1.3mm}{\scalebox{.5}{RagInit}}} & 
            \includegraphics[width=0.15\columnwidth, frame]{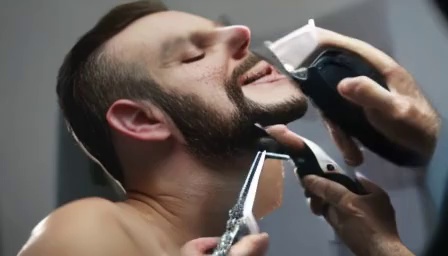} &
            \includegraphics[width=0.15\columnwidth, frame]{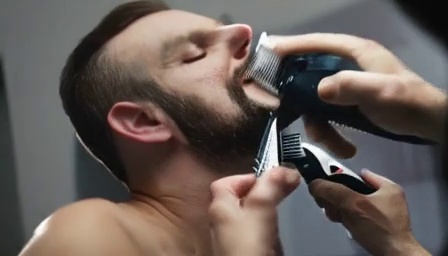} &
            \includegraphics[width=0.15\columnwidth, frame]{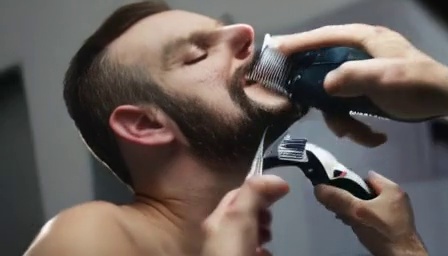} &
            \includegraphics[width=0.15\columnwidth, frame]{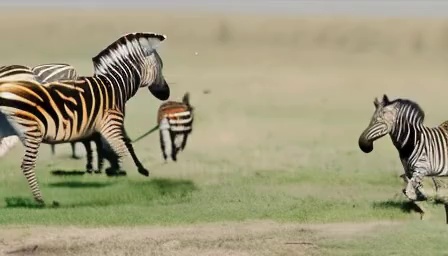} &
            \includegraphics[width=0.15\columnwidth, frame]{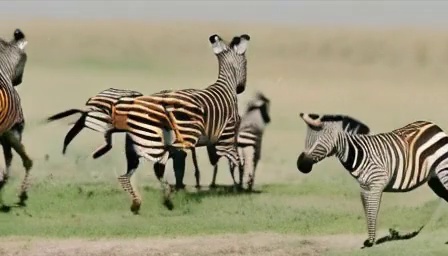} &
            \includegraphics[width=0.15\columnwidth, frame]{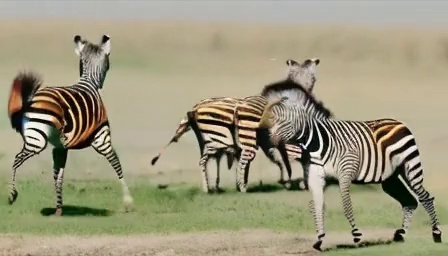}  \\

            \rotatebox{90}{\hspace{1.3mm}{\scalebox{.5}{\textbf{\methodname}}}} & 
            \includegraphics[width=0.15\columnwidth, frame]{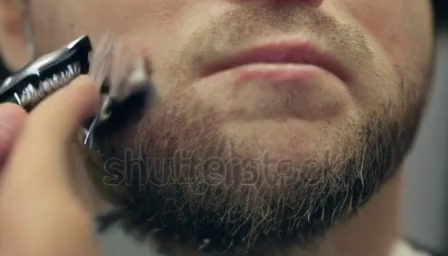} &
            \includegraphics[width=0.15\columnwidth, frame]{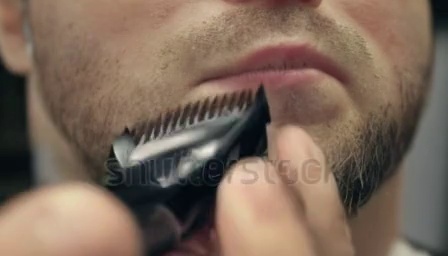} &
            \includegraphics[width=0.15\columnwidth, frame]{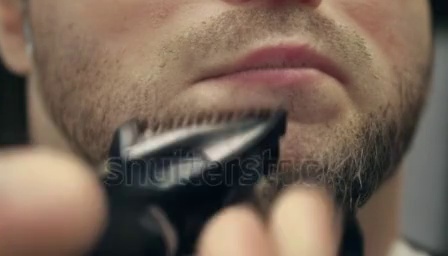} &
            \includegraphics[width=0.15\columnwidth, frame]{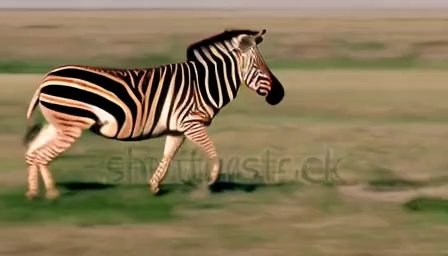} &
            \includegraphics[width=0.15\columnwidth, frame]{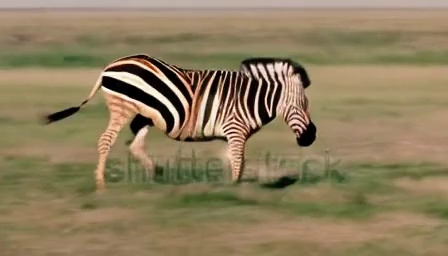} &
            \includegraphics[width=0.15\columnwidth, frame]{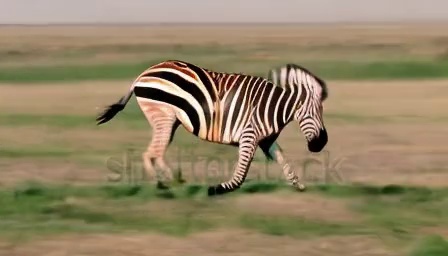} \\
            & \multicolumn{3}{c}{\scalebox{.5}{\texttt{"A person is cutting or shaving the beard."}}} & \multicolumn{3}{c}{\scalebox{0.5}{\texttt{"A zebra running to join a herd of its kind."}}} \\

            \rotatebox{90}{\hspace{0.5mm}{\scalebox{.5}{ZeroScope}}} & 
            \includegraphics[width=0.15\columnwidth, frame]{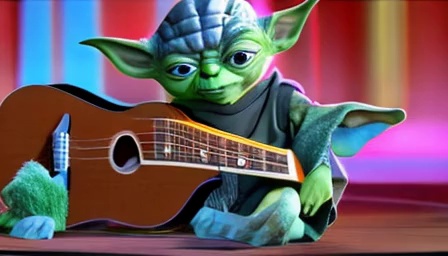} &
            \includegraphics[width=0.15\columnwidth, frame]{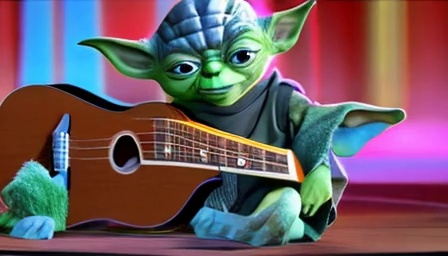} &
            \includegraphics[width=0.15\columnwidth, frame]{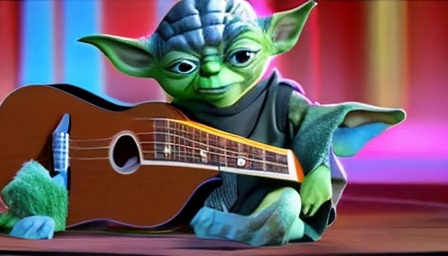} &
            \includegraphics[width=0.15\columnwidth, frame]{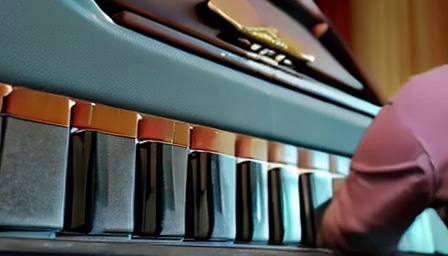} &
            \includegraphics[width=0.15\columnwidth, frame]{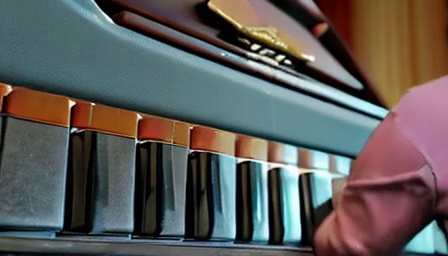} &
            \includegraphics[width=0.15\columnwidth, frame]{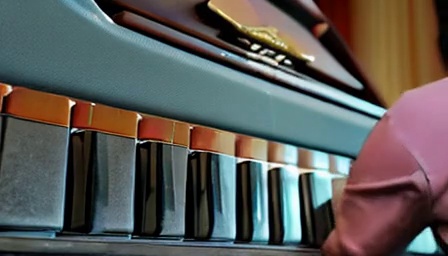}  \\

            \rotatebox{90}{\hspace{0.8mm}{\scalebox{.5}{FreeInit}}} & 
            \includegraphics[width=0.15\columnwidth, frame]{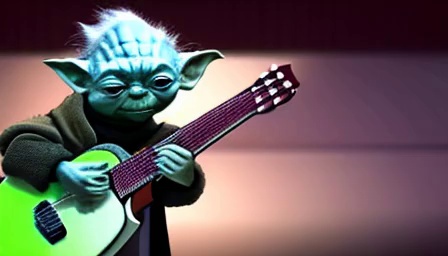} &
            \includegraphics[width=0.15\columnwidth, frame]{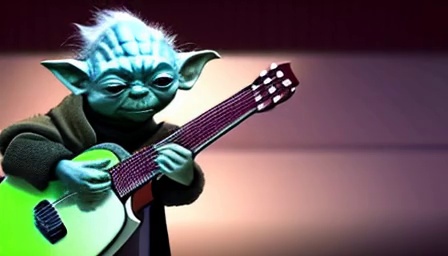} &
            \includegraphics[width=0.15\columnwidth, frame]{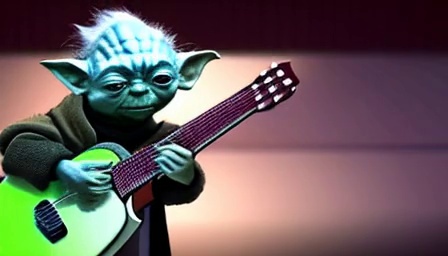} &
            \includegraphics[width=0.15\columnwidth, frame]{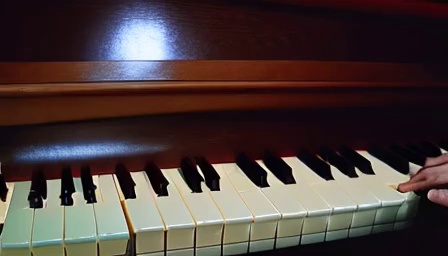} &
            \includegraphics[width=0.15\columnwidth, frame]{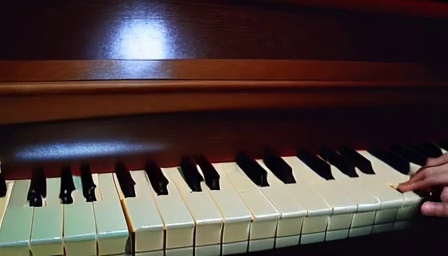} &
            \includegraphics[width=0.15\columnwidth, frame]{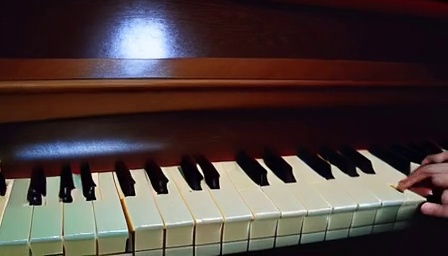}  \\

            \rotatebox{90}{\hspace{1.3mm}{\scalebox{.5}{RagInit}}} & 
            \includegraphics[width=0.15\columnwidth, frame]{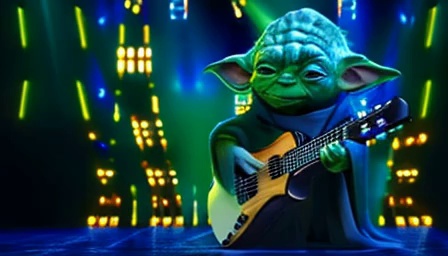} &
            \includegraphics[width=0.15\columnwidth, frame]{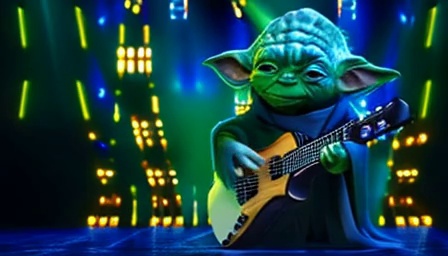} &
            \includegraphics[width=0.15\columnwidth, frame]{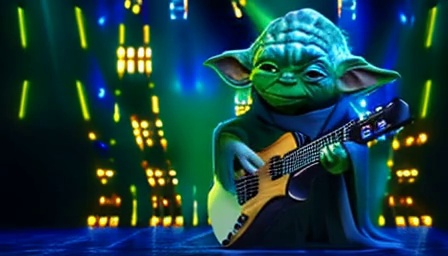} &
            \includegraphics[width=0.15\columnwidth, frame]{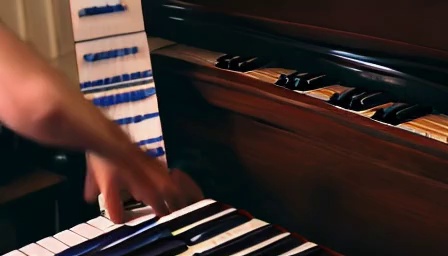} &
            \includegraphics[width=0.15\columnwidth, frame]{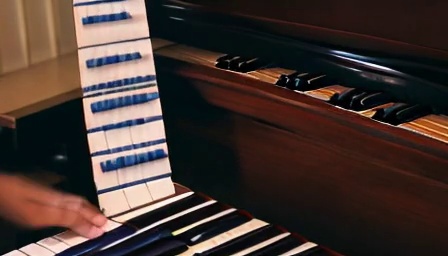} &
            \includegraphics[width=0.15\columnwidth, frame]{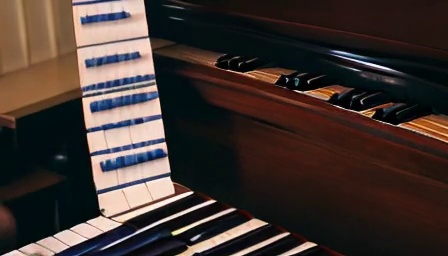}  \\

            \rotatebox{90}{\hspace{1.3mm}{\scalebox{.5}{\textbf{\methodname}}}} & 
            \includegraphics[width=0.15\columnwidth, frame]{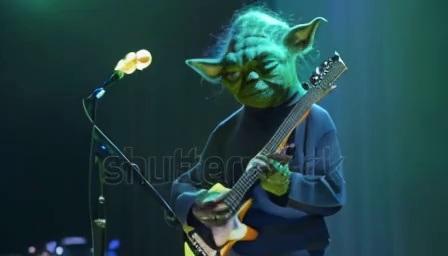} &
            \includegraphics[width=0.15\columnwidth, frame]{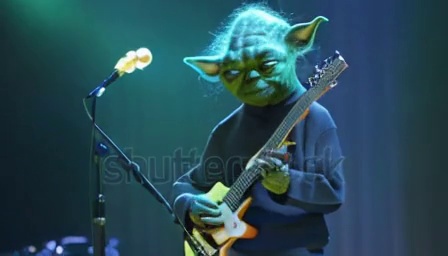} &
            \includegraphics[width=0.15\columnwidth, frame]{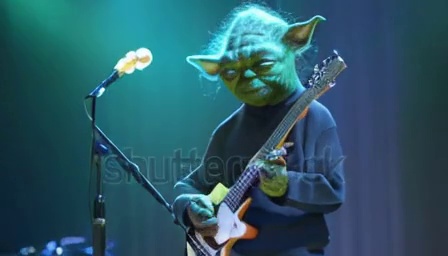} &
            \includegraphics[width=0.15\columnwidth, frame]{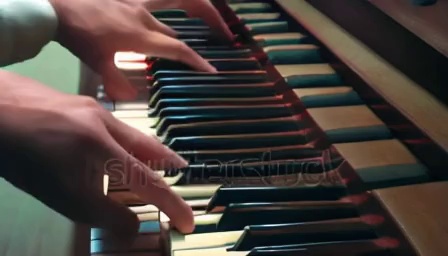} &
            \includegraphics[width=0.15\columnwidth, frame]{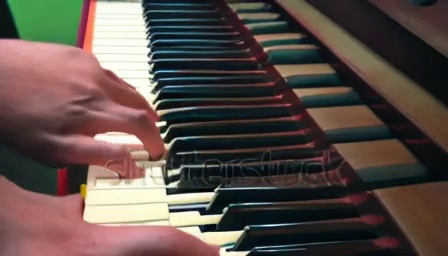} &
            \includegraphics[width=0.15\columnwidth, frame]{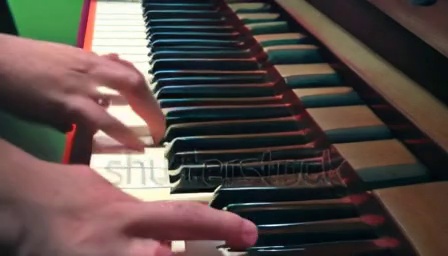} \\
            & \multicolumn{3}{c}{\scalebox{.5}{\texttt{"Yoda playing guitar on the stage."}}} & \multicolumn{3}{c}{\scalebox{0.5}{\texttt{"A person is playing piano."}}}
    \end{tabular}
    }
    \captionof{figure}{Visual comparison of the different methods. We report the prompt at the bottom.}
    \label{fig:comparison}
\end{table*}

\paragraph{Number of retrieved examples $K$} We study the impact of the number of retrieved samples on the final generated videos, comparing the FVD vs DINO-similarity trade-off. Specifically, we train different versions of the models to use different numbers of $K$. We use a reduced computation budget and train the models for $25k$ iterations. We report the results in \cref{fig:role_of_k}. We can observe that $K=1$, \ie retrieving a single sample, achieves good FVD but incurs very high DINO-similarity (\ie undesired copy-paste-effects). Vice-versa, increasing $K$ too much, results in progressively worse FVD probably because it becomes harder for the model to get meaningful information (besides incurring additional computational costs). We observe that $K=5$ represents a good trade-off. We set this value and use it throughout all our experiments. In principle, nothing prevents us from training a model with a given $K$ and adopting a different $K'$  at inference time. However, we observed slightly reduced performances.
We add a more detailed discussion, along with qualitative results for different $K$ in the supplementary material. 

\paragraph{Computational Complexity} Lastly we discuss the computational complexity added by our method. Running a Diffusion Model is computationally expensive, mainly due to the cost of the denoiser network. 
However, the main computational burden of \methodname{} is encoding the retrieved videos with CLIP and the VAE encoder to obtain the latent for the initialization. All these steps can be easily parallelized and introduce negligible computation and latency, while the retrieval is high-speed thanks to the FAISS library \cite{douze2024faiss}. In total, this amounts to an increased latency of 20\% to generate a single video.

\begin{table*}[!th]
    \centering
    \resizebox{\textwidth}{!}{
    \setlength\tabcolsep{0.3pt}
    \footnotesize
    \renewcommand{\arraystretch}{0.3}
    \begin{tabular}{c@{\hskip 1.5pt}ccccc}
            \scalebox{0.5}{Generated} & \multicolumn{5}{c}{\scalebox{0.5}{Retrieved Samples}} \\
            \includegraphics[width=0.13\columnwidth, frame]{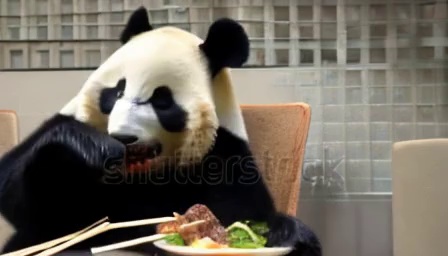} &
            \includegraphics[width=0.13\columnwidth, frame]{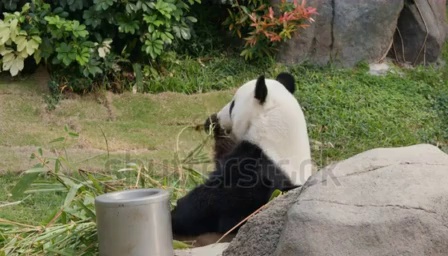} &
            \includegraphics[width=0.13\columnwidth, frame]{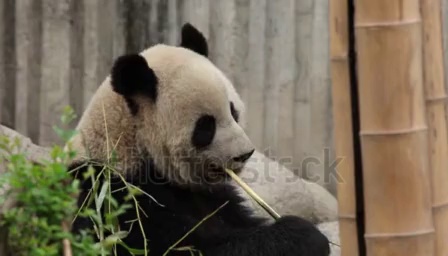} &
            \includegraphics[width=0.13\columnwidth, frame]{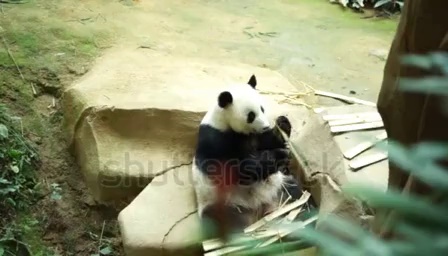} &
            \includegraphics[width=0.13\columnwidth, frame]{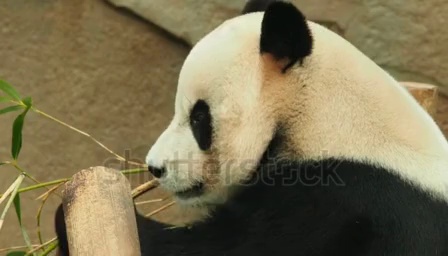} &
            \includegraphics[width=0.13\columnwidth, frame]{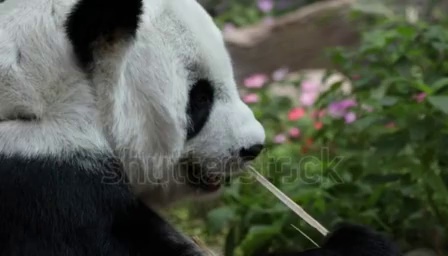}  \\
            \multicolumn{6}{c}{\scalebox{0.5}{\texttt{"A cute panda eating Chinese food in a restaurant."}}} \\

            \includegraphics[width=0.13\columnwidth, frame]{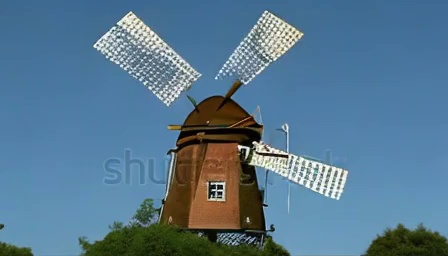} &
            \includegraphics[width=0.13\columnwidth, frame]{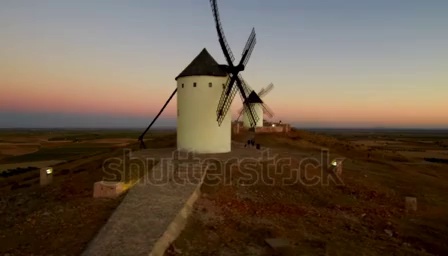} &
            \includegraphics[width=0.13\columnwidth, frame]{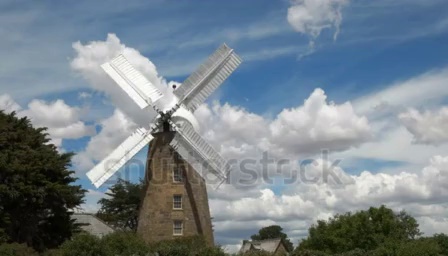} &
            \includegraphics[width=0.13\columnwidth, frame]{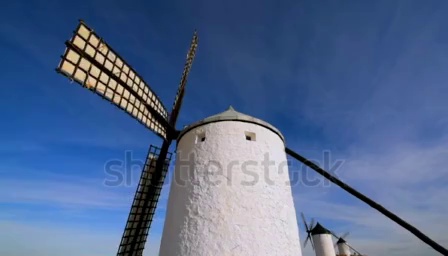} &
            \includegraphics[width=0.13\columnwidth, frame]{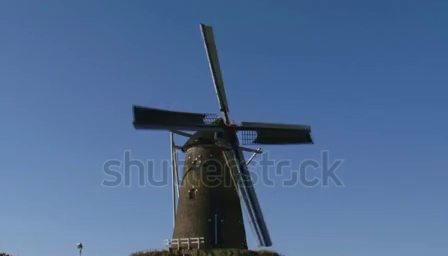} &
            \includegraphics[width=0.13\columnwidth, frame]{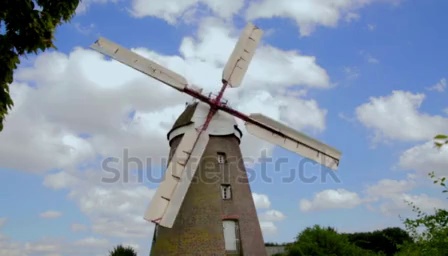}  \\
            \multicolumn{6}{c}{\scalebox{0.5}{\texttt{"A windmill."}}} \\

            \includegraphics[width=0.13\columnwidth, frame]{figures/comparison/our/005/0000.jpg} &
            \includegraphics[width=0.13\columnwidth, frame]{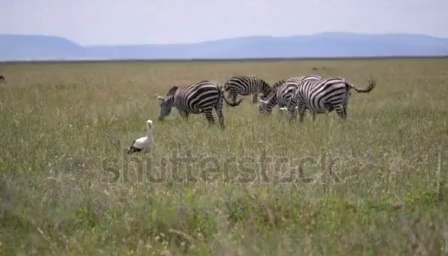} &
            \includegraphics[width=0.13\columnwidth, frame]{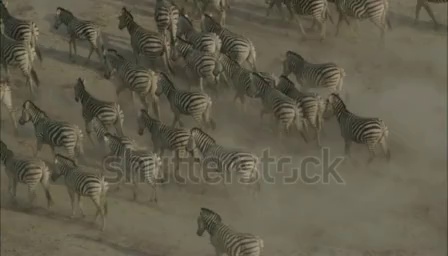} &
            \includegraphics[width=0.13\columnwidth, frame]{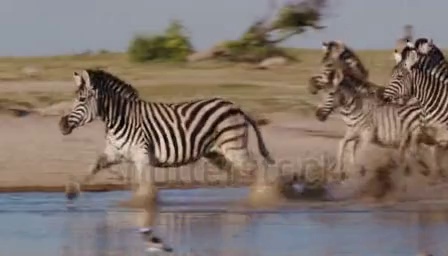} &
            \includegraphics[width=0.13\columnwidth, frame]{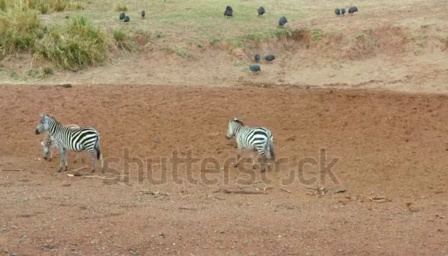} &
            \includegraphics[width=0.13\columnwidth, frame]{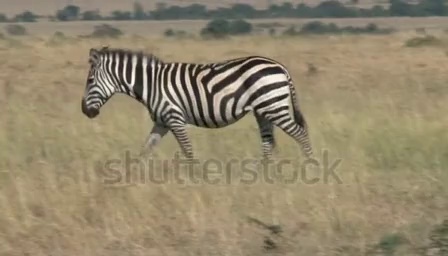}  \\
            \multicolumn{6}{c}{\scalebox{0.5}{\texttt{"A zebra running to join a herd of its kind"}}} \\


    \end{tabular}
    }
    \captionof{figure}{We show the first frame of the generated video and the first frame of the 5 retrieved samples used during the generation phase. No clear leakage is present, \ie the model is not simply copy-pasting the output but using it to improve the result.}
    \label{fig:retrived}
\end{table*}

\subsection{Qualitative Results}
In this section, we present a qualitative comparison between different methods, moreover, we explore an additional use case of our method \ie \emph{motion transfer}. In \cref{fig:comparison}, we display frames from the generated videos based on prompts from the VBench suite. Our methods produce better videos in terms of both motion and scene composition. Additionally, in \cref{fig:retrived}, we show the first frame of the generated video alongside the first frames of the five videos used for conditioning. We observe that no clear leakage is present, indicating that \methodname{} effectively integrates the retrieved information to achieve better results. The generated videos from our method contain watermarks due to the training dataset, WebVid10M \cite{Bain21}. However, training on higher-quality datasets would eliminate this artifact.

\section{Conclusions}
\label{sec:conclusions}
In this work, we propose \methodname{} a framework for retrieval augmented text-to-video generation. We exploit retrieved videos to enhance the motion realism of the final result, showing superior performance both qualitatively and quantitatively. Moreover, we showcase how this framework can be adapted to specific tasks such as Motion Transfer, obtaining results on par with state-of-the-art at a fraction of the computational costs.

Our work opens the door to several future improvements. First, exploring the use of alternative encoders, such as video models, could provide more robust representations of actions. 
Extending our approach to other diffusion models and transformer-based architectures could further generalize the method, making it suitable for a wider range of applications. Lastly, expanding the model to handle the composition of multiple actions—rather than assuming a single action—would also broaden its applicability. 



{
    \small
    \bibliographystyle{ieeenat_fullname}
    \bibliography{main}
}

\clearpage
\setcounter{page}{1}
\maketitlesupplementary
\appendix

We provide additional details and results for our method. In \cref{sec:implementation}, we delve deeper into the implementation of \methodname{}, with a particular focus on the retrieval system. Following this, we present both qualitative and quantitative results. In \cref{sec:metrics}, we report the full evaluation metrics on the VBench suite. Lastly, in \cref{sec:qualitative} we showcase additional qualitative results. 

\section{Implementation}
\label{sec:implementation}

We provide additional details on the implementation of our retrieval mechanism. We build our retrieval system on the WebVid10M [1] dataset. First, we use the CLIP ViT-B/32 model to encode the video frames. This model includes both image and text encoders, which produce embeddings of size $\text{dim} = 512$. Next, we leverage the FAISS library [13] to create an index for efficient retrieval. The WebVid10M dataset contains duplicate or highly similar videos; to prevent the model from processing redundant information, given a query $\mathbf{q}$, we apply a deduplication strategy based on the cosine similarity between samples. We empirically set the deduplication threshold at $\delta_{\text{dedup}} = 0.965$ and maintain this value across all experiments. Additionally, to ensure that the retrieved videos are relevant to the query, we set a minimum cosine similarity threshold of $\delta_{\min} = 0.6$ and remove samples from the retrieval set $\mathbf{Z}$ that do not meet this criterion. 
In such cases, padding is used to match the required length.

From an architectural point of view, we introduce the transformer temporal enhancer module to improve the temporal representation of the video. It is composed of 6 layers of transformer blocks with a hidden dimension of $\text{dim} = 512$. A learnable token $\texttt{[CLS]}$ is added at the beginning of the sequence and pooled in output to represent the video. Lastly, we add multi-head cross-attention layers to the base T2V model ZeroScope. We introduce a point-wise convolution initialized with zero-weights, to act as the identity when the model is initialized. 

The added modules are finetuned (while keeping the rest of the network frozen) for 200K iterations on the WebVid10M dataset, at resolution $448 \times 256$ and $12$ frames. Training is performed with an effective batch size of 16, distributed on 4 Nvidia A100 GPUs.

\section{VBench Results}
\label{sec:metrics}
\begin{figure}
    \vspace{-2.5em}
    \centering
    \includegraphics[width=\linewidth]{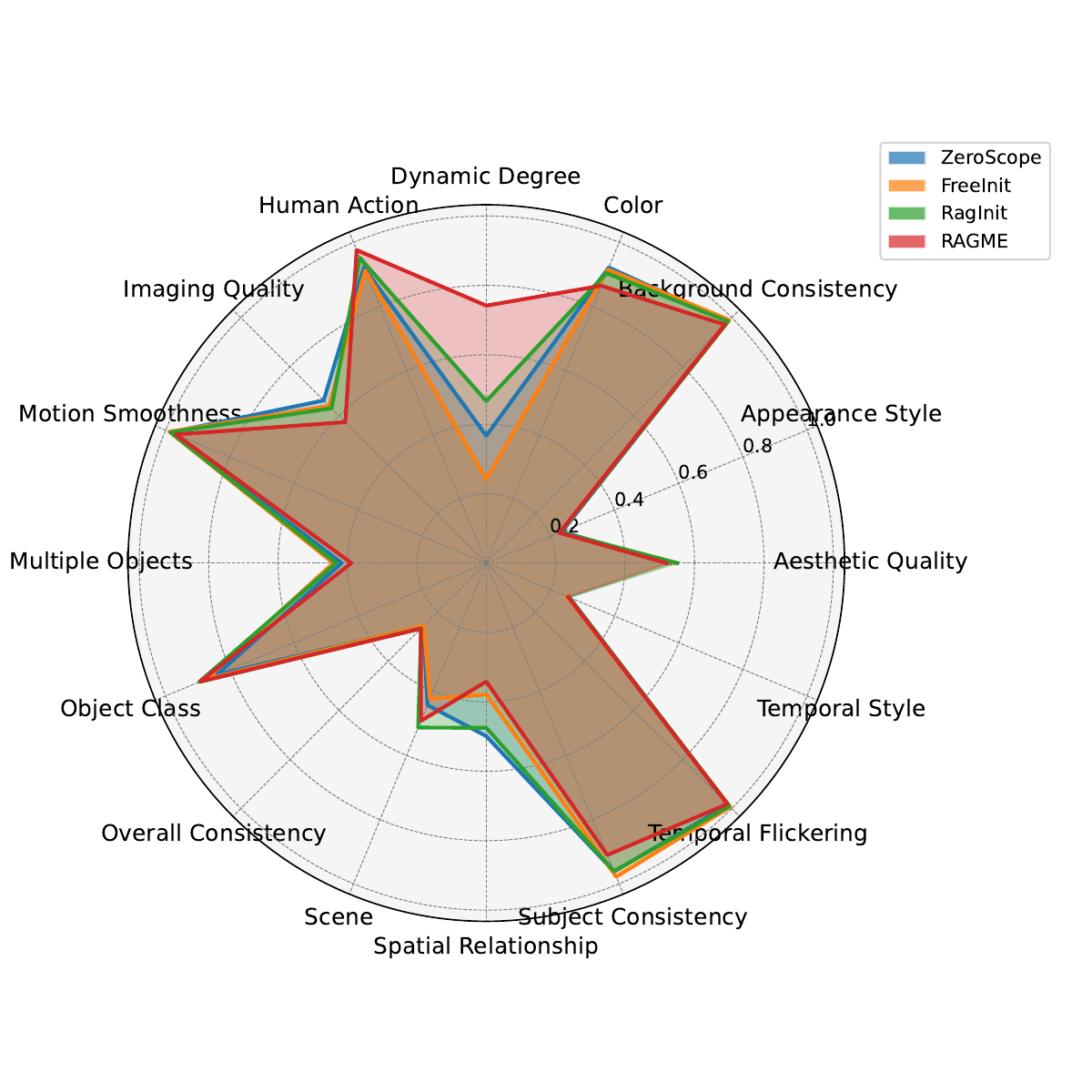}
    \vspace{-2em}
    \caption{Full comparison on the VBench benchmark.}
    \label{fig:vbench}
\end{figure}

We report all the metrics from the VBench benchmark in \cref{fig:vbench}, which complements the results of Tab. 2 of the main paper. We can observe that the methods perform similarly on many metrics, with some noticeable exceptions. \methodname{} outperforms the baseline on the motion-related metrics (\eg Dynamic Degree and Human Action), while falling short on Image Quality and Subject Consistency. The first can be explained by the low quality of the WebVid10M dataset (\eg, the presence of the watermark) which can deteriorate the quality of the generated frames. The second is linked with the increased motion, which would inevitably make the consistency harder. However, from visual inspection, we didn't notice a significant drop in the quality of the videos nor temporal artifacts such as flickering or inconsistent objects. 

\section{Qualitative Results}
\label{sec:qualitative}
In \cref{fig:comparison_supp}, we present additional videos for the VBench prompts. \methodname{} generates better results also in the presence of complex or objects prompt (\eg the last row).  Next, we compare the first frame of the generated video with the first frame of the retrieved samples, showing that the model is not directly coping with the conditioning signal.

\begin{table*}[t]
    \centering
    \resizebox{\textwidth}{!}{
    \setlength\tabcolsep{0.3pt}
    \footnotesize
    \renewcommand{\arraystretch}{0.3}
    \begin{tabular}{cccc}
            \rotatebox{90}{\hspace{0.5mm}{\scalebox{.5}{Ref. Video}}} & 
            \includegraphics[width=0.20\columnwidth, frame]{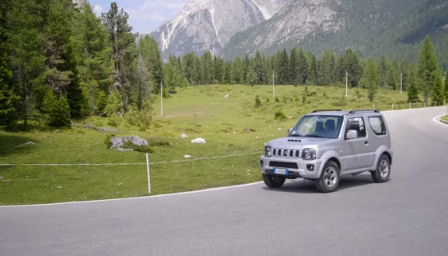} &
            \includegraphics[width=0.20\columnwidth, frame]{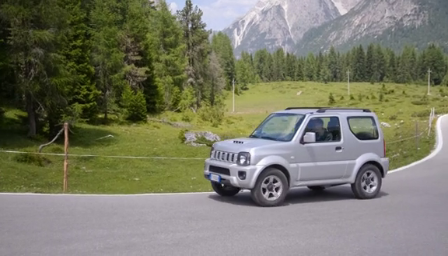} &
            \includegraphics[width=0.20\columnwidth, frame]{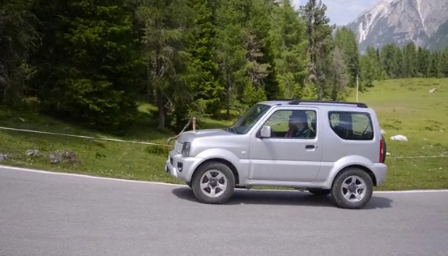}  \\

            \rotatebox{90}{\hspace{1.0mm}{\scalebox{.5}{MD \cite{zhao2024motiondirector}}}} & 
            \includegraphics[width=0.20\columnwidth, frame]{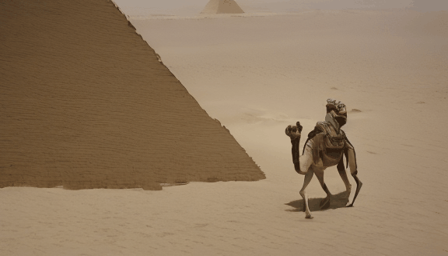} &
            \includegraphics[width=0.20\columnwidth, frame]{figures/motion_customization/md/person/frame_000.png} &
            \includegraphics[width=0.20\columnwidth, frame]{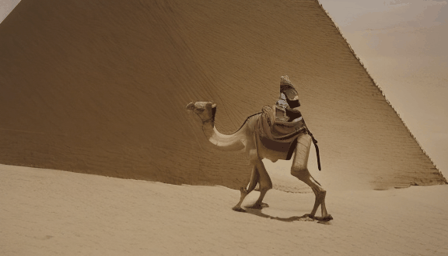}  \\

            \rotatebox{90}{\hspace{1.0mm}{\scalebox{.5}{\textbf{\methodname}}}} & 
            \includegraphics[width=0.20\columnwidth, frame]{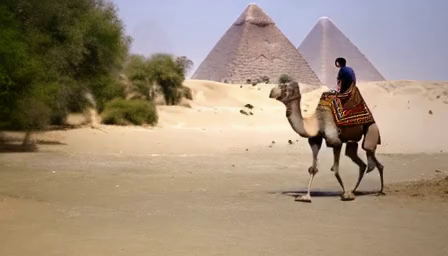} &
            \includegraphics[width=0.20\columnwidth, frame]{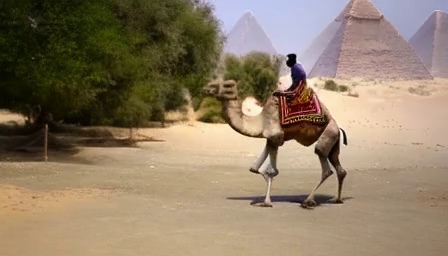} &
            \includegraphics[width=0.20\columnwidth, frame]{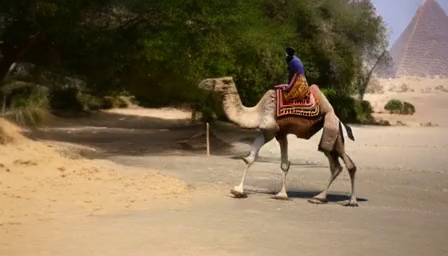}  \\ 
            \multicolumn{4}{c}{\scalebox{0.5}{\texttt{``A person is riding a camel with the Pyramids as background.''}}}\\

            \rotatebox{90}{\hspace{1.0mm}{\scalebox{.5}{MD \cite{zhao2024motiondirector}}}} & 
            \includegraphics[width=0.20\columnwidth, frame]{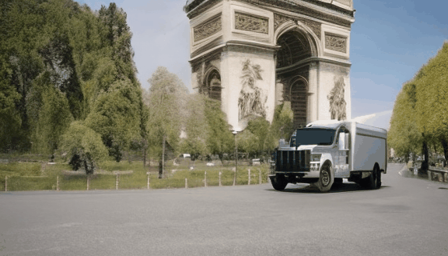} &
            \includegraphics[width=0.20\columnwidth, frame]{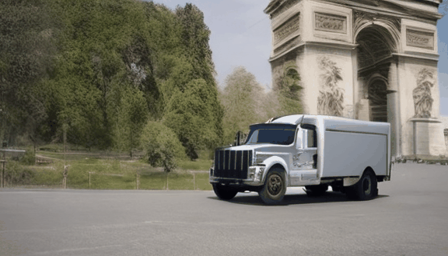} &
            \includegraphics[width=0.20\columnwidth, frame]{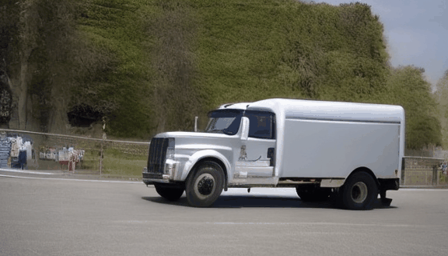}  \\

            \rotatebox{90}{\hspace{1.0mm}{\scalebox{.5}{\textbf{\methodname}}}} & 
            \includegraphics[width=0.20\columnwidth, frame]{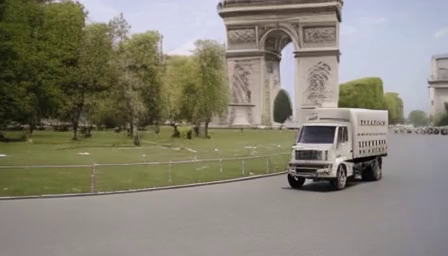} &
            \includegraphics[width=0.20\columnwidth, frame]{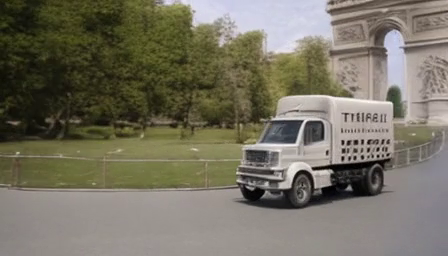} &
            \includegraphics[width=0.20\columnwidth, frame]{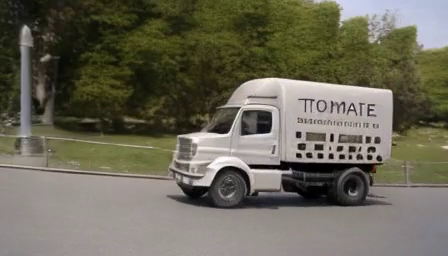}  \\
            \multicolumn{4}{c}{\scalebox{0.5}{\texttt{``A truck is driving past the Arc de Triomphe.''}}}
    \end{tabular}
    }
    \captionof{figure}{Results for the motion transfer task. The top row displays the reference video, followed by a comparison of Motion Director (MD) \cite{zhao2024motiondirector} and our method using two distinct prompts (shown at the bottom). Our approach achieves qualitatively similar results with 8$\times$ fewer fine-tuning iterations.}
    \label{fig:motion_customization}
\end{table*}

\paragraph{Motion Transfer} While our method is designed for flexible conditioning on multiple retrieved videos, a key application in video editing is motion transfer. This involves transferring motion from a reference video while controlling the appearance and overall style of the output, for instance, through a textual prompt. 

Our approach is specifically designed to avoid explicit copy-paste artifacts, extracting only high-level motion semantics from the retrieved videos - aligning with our goal of enhancing generated motion in a generalizable way. However, for motion transfer, we can adapt our method accordingly. In practice, given a driving video, we overfit the controller network to that specific video to achieve the desired effect. Notably, the design of our architecture and pretraining on WebVid-10M facilitate this adaptation process, making it more efficient compared to other methods that require fine-tuning on the target video.  Compared to Motion Director \cite{zhao2024motiondirector} (which relies on the same backbone video generator), our method achieves similar performance while requiring eight times less fine-tuning (50 vs 400 iterations), demonstrating the efficiency of our RAG-first design.

\begin{table*}
    \centering
    \resizebox{\textwidth}{!}{
    \setlength\tabcolsep{0.3pt}
    \footnotesize
    \renewcommand{\arraystretch}{0.3}
    \begin{tabular}{cccc@{\hskip 1pt}ccc}
            \rotatebox{90}{\hspace{0.5mm}{\scalebox{.5}{ZeroScope}}} & 
            \includegraphics[width=0.18\columnwidth, frame]{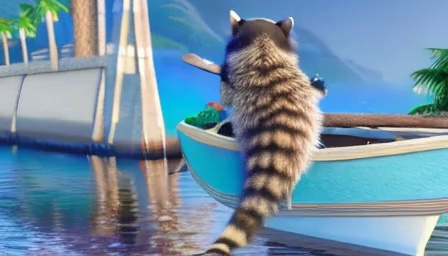} &
            \includegraphics[width=0.18\columnwidth, frame]{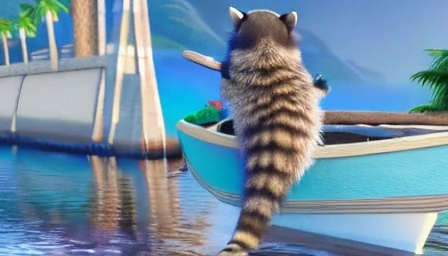} &
            \includegraphics[width=0.18\columnwidth, frame]{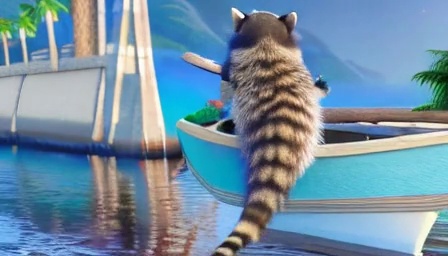} &
            \includegraphics[width=0.18\columnwidth, frame]{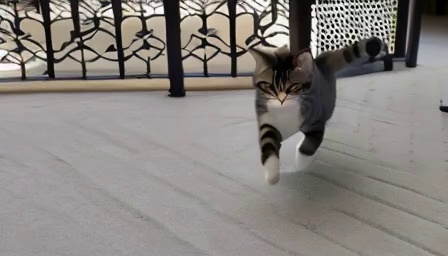} &
            \includegraphics[width=0.18\columnwidth, frame]{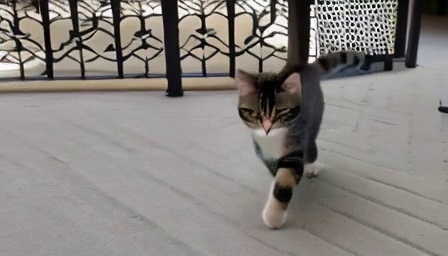} &
            \includegraphics[width=0.18\columnwidth, frame]{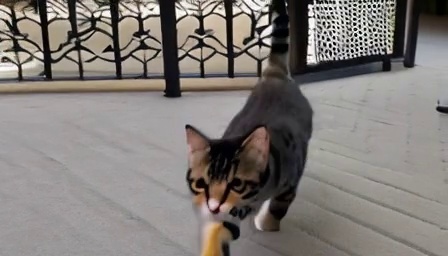}  \\

            \rotatebox{90}{\hspace{0.8mm}{\scalebox{.5}{FreeInit}}} & 
            \includegraphics[width=0.18\columnwidth, frame]{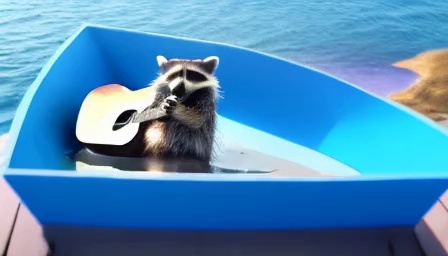} &
            \includegraphics[width=0.18\columnwidth, frame]{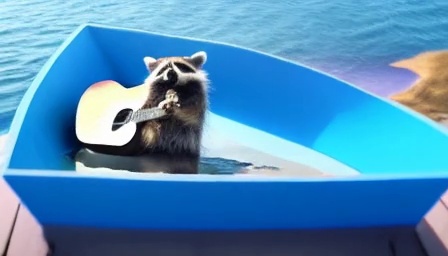} &
            \includegraphics[width=0.18\columnwidth, frame]{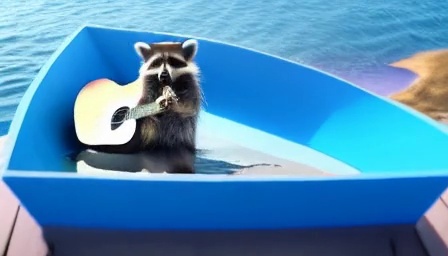} &
            \includegraphics[width=0.18\columnwidth, frame]{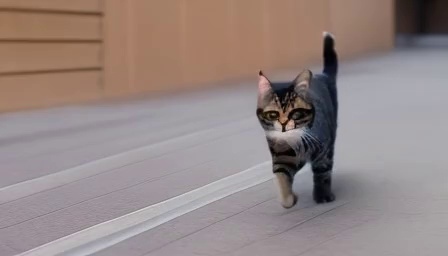} &
            \includegraphics[width=0.18\columnwidth, frame]{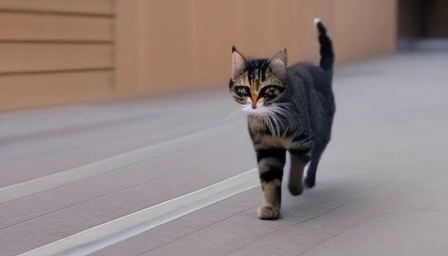} &
            \includegraphics[width=0.18\columnwidth, frame]{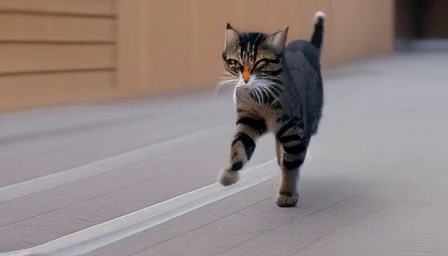}  \\

            \rotatebox{90}{\hspace{2.0mm}{\scalebox{.5}{RagInit}}} & 
            \includegraphics[width=0.18\columnwidth, frame]{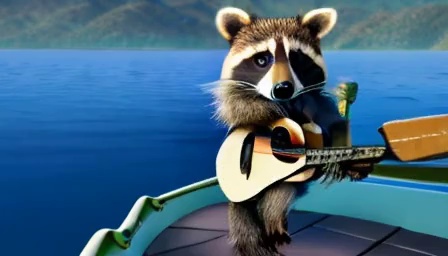} &
            \includegraphics[width=0.18\columnwidth, frame]{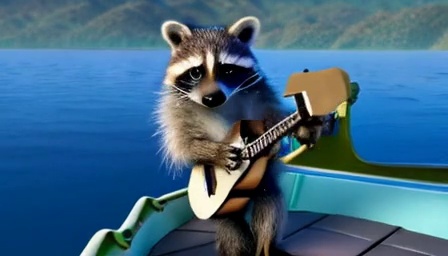} &
            \includegraphics[width=0.18\columnwidth, frame]{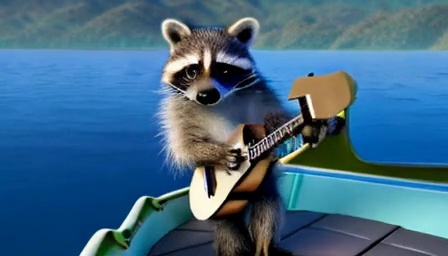} &
            \includegraphics[width=0.18\columnwidth, frame]{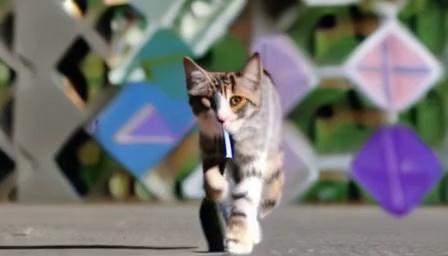} &
            \includegraphics[width=0.18\columnwidth, frame]{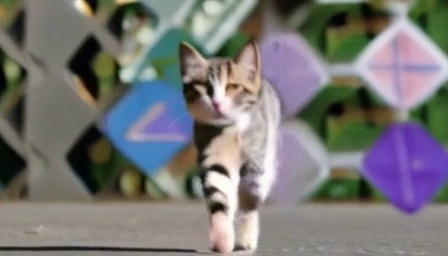} &
            \includegraphics[width=0.18\columnwidth, frame]{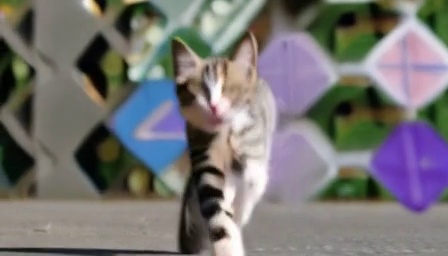}  \\

            \rotatebox{90}{\hspace{2.0mm}{\scalebox{.5}{\textbf{\methodname}}}} & 
            \includegraphics[width=0.18\columnwidth, frame]{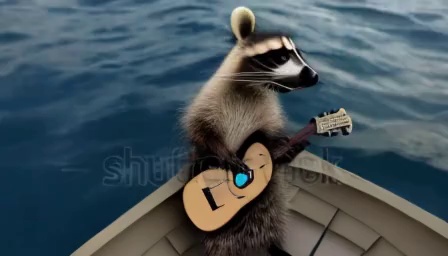} &
            \includegraphics[width=0.18\columnwidth, frame]{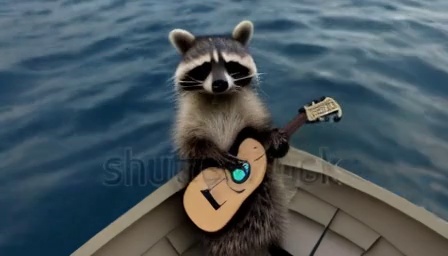} &
            \includegraphics[width=0.18\columnwidth, frame]{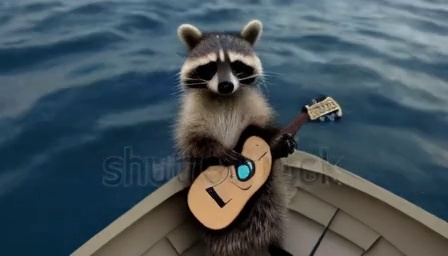} &
            \includegraphics[width=0.18\columnwidth, frame]{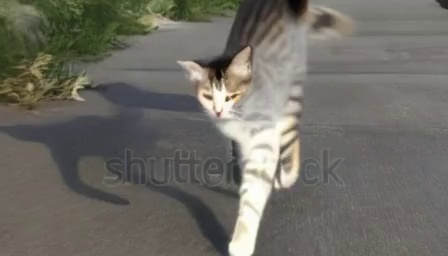} &
            \includegraphics[width=0.18\columnwidth, frame]{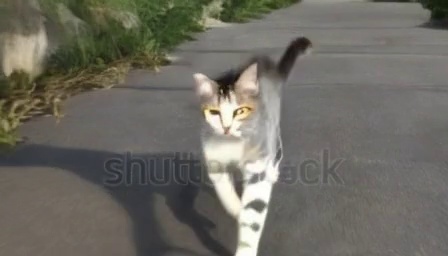} &
            \includegraphics[width=0.18\columnwidth, frame]{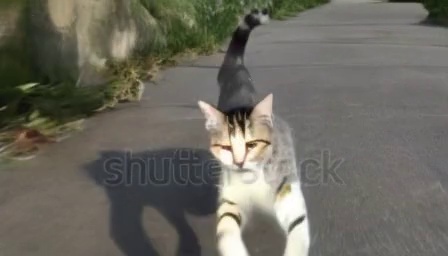} \\
            & \multicolumn{3}{c}{\scalebox{.5}{\texttt{"A cute raccoon playing guitar in a boat on the ocean."}}} & \multicolumn{3}{c}{\scalebox{0.5}{\texttt{"A cat running happily."}}} \\\\

                        \rotatebox{90}{\hspace{0.5mm}{\scalebox{.5}{ZeroScope}}} & 
            \includegraphics[width=0.18\columnwidth, frame]{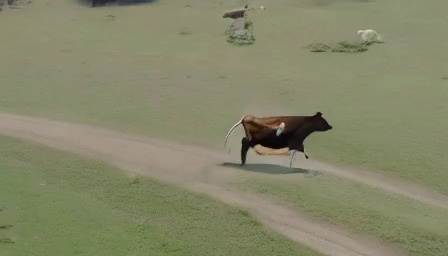} &
            \includegraphics[width=0.18\columnwidth, frame]{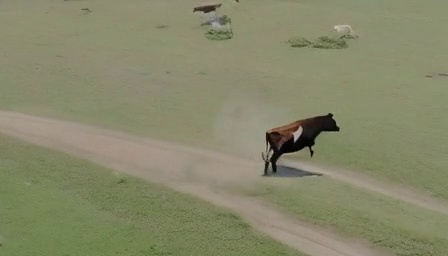} &
            \includegraphics[width=0.18\columnwidth, frame]{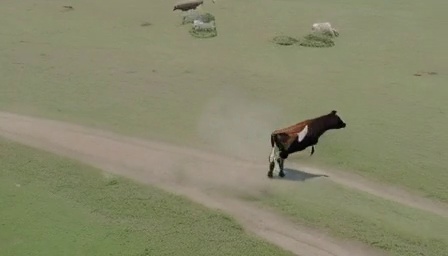} &
            \includegraphics[width=0.18\columnwidth, frame]{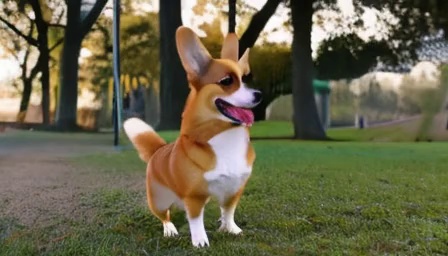} &
            \includegraphics[width=0.18\columnwidth, frame]{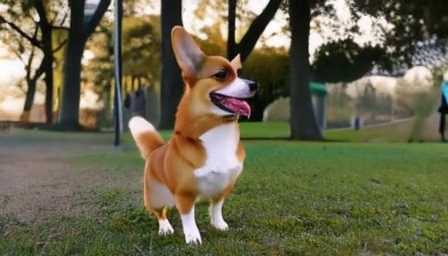} &
            \includegraphics[width=0.18\columnwidth, frame]{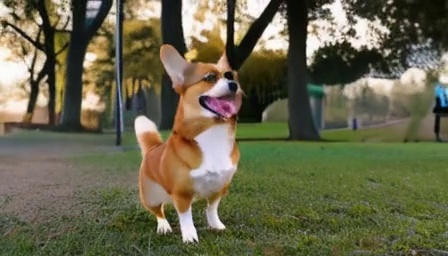}  \\

            \rotatebox{90}{\hspace{0.8mm}{\scalebox{.5}{FreeInit}}} & 
            \includegraphics[width=0.18\columnwidth, frame]{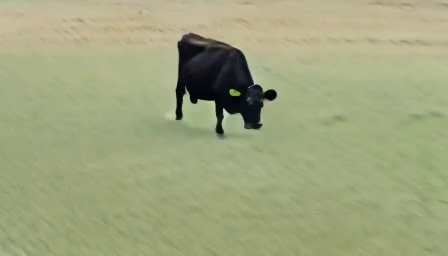} &
            \includegraphics[width=0.18\columnwidth, frame]{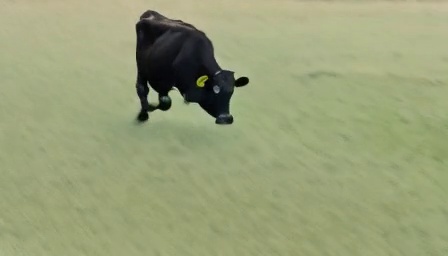} &
            \includegraphics[width=0.18\columnwidth, frame]{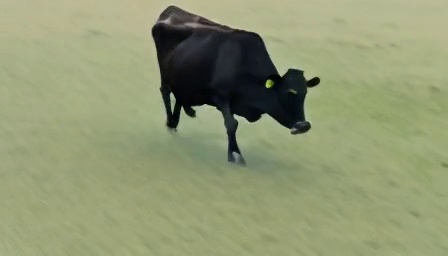} &
            \includegraphics[width=0.18\columnwidth, frame]{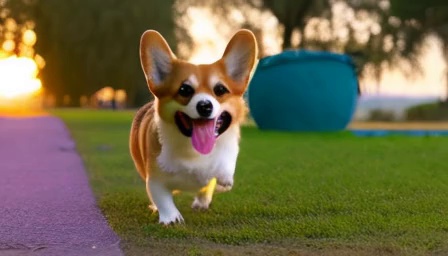} &
            \includegraphics[width=0.18\columnwidth, frame]{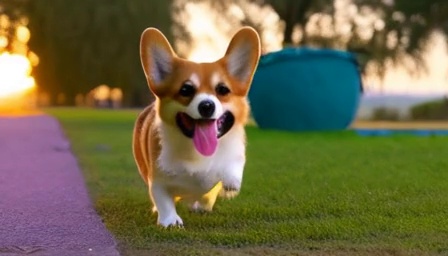} &
            \includegraphics[width=0.18\columnwidth, frame]{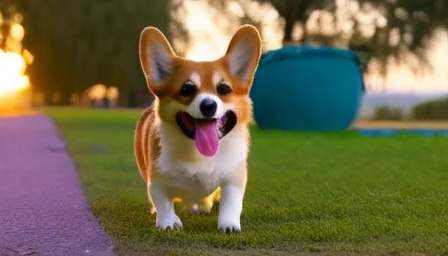}  \\

            \rotatebox{90}{\hspace{2.0mm}{\scalebox{.5}{RagInit}}} & 
            \includegraphics[width=0.18\columnwidth, frame]{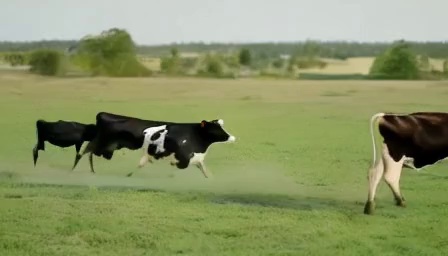} &
            \includegraphics[width=0.18\columnwidth, frame]{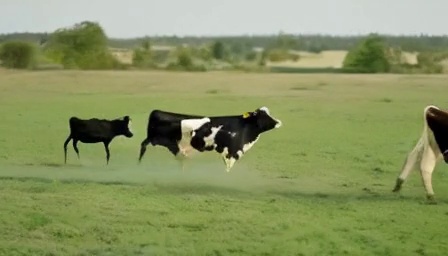} &
            \includegraphics[width=0.18\columnwidth, frame]{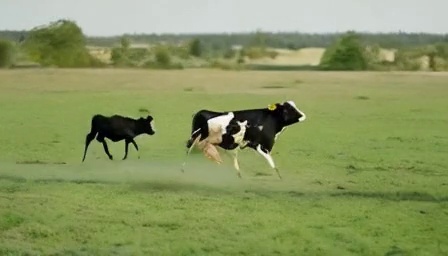} &
            \includegraphics[width=0.18\columnwidth, frame]{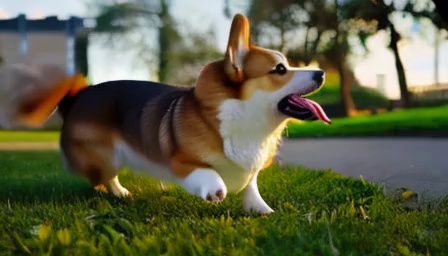} &
            \includegraphics[width=0.18\columnwidth, frame]{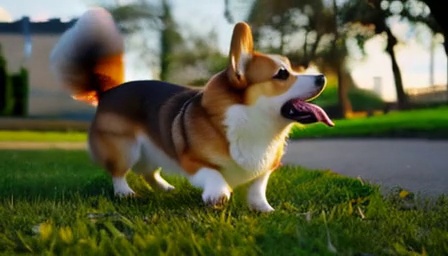} &
            \includegraphics[width=0.18\columnwidth, frame]{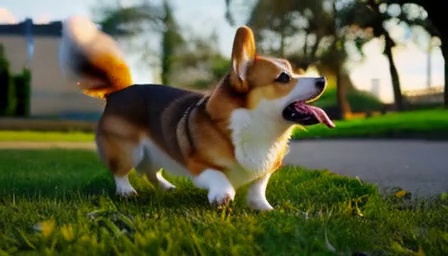}  \\

            \rotatebox{90}{\hspace{2.0mm}{\scalebox{.5}{\textbf{\methodname}}}} & 
            \includegraphics[width=0.18\columnwidth, frame]{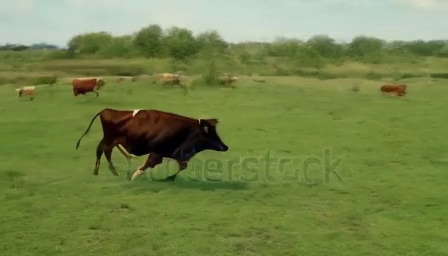} &
            \includegraphics[width=0.18\columnwidth, frame]{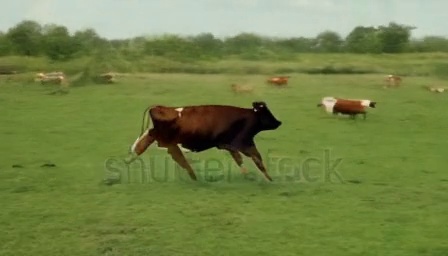} &
            \includegraphics[width=0.18\columnwidth, frame]{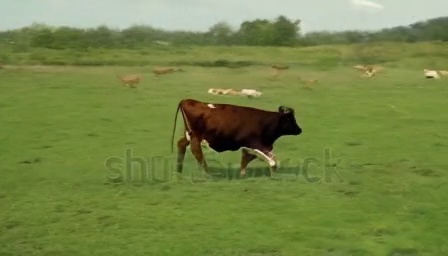} &
            \includegraphics[width=0.18\columnwidth, frame]{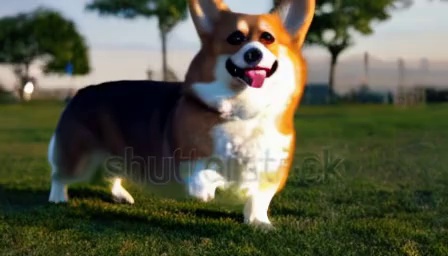} &
            \includegraphics[width=0.18\columnwidth, frame]{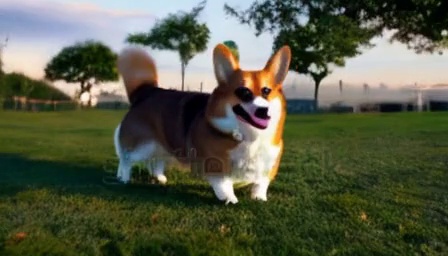} &
            \includegraphics[width=0.18\columnwidth, frame]{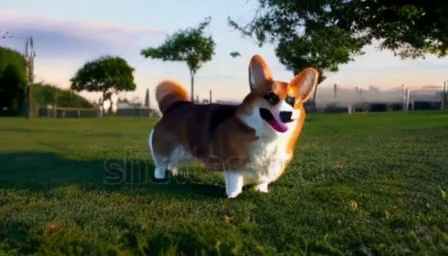} \\
            & \multicolumn{3}{c}{\scalebox{.5}{\texttt{"A cow running to join a herd of its kind."}}} & \multicolumn{3}{c}{\scalebox{0.5}{\texttt{"A cute happy Corgi playing in park, sunset, zoom out."}}} \\\\

            \rotatebox{90}{\hspace{0.5mm}{\scalebox{.5}{ZeroScope}}} & 
            \includegraphics[width=0.18\columnwidth, frame]{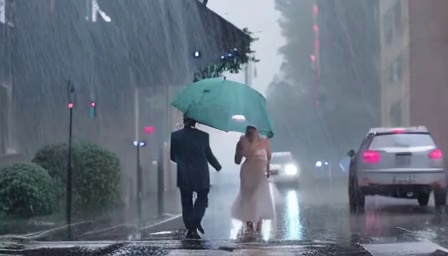} &
            \includegraphics[width=0.18\columnwidth, frame]{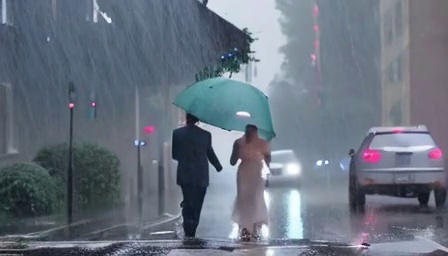} &
            \includegraphics[width=0.18\columnwidth, frame]{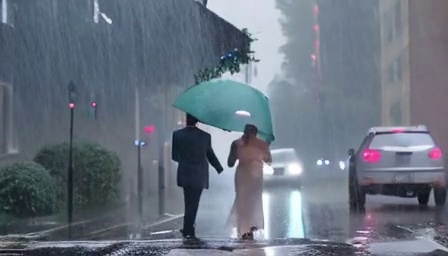} &
            \includegraphics[width=0.18\columnwidth, frame]{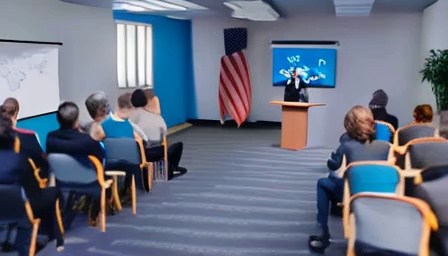} &
            \includegraphics[width=0.18\columnwidth, frame]{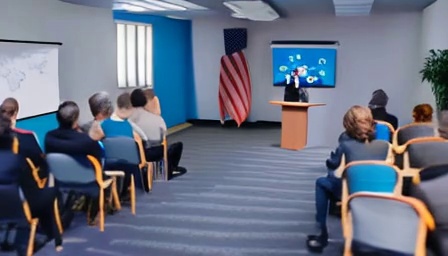} &
            \includegraphics[width=0.18\columnwidth, frame]{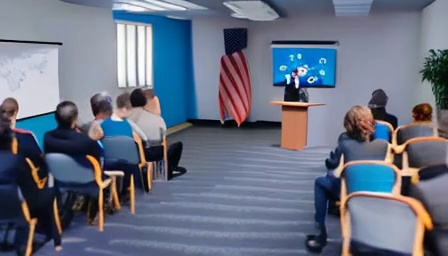}  \\

            \rotatebox{90}{\hspace{0.8mm}{\scalebox{.5}{FreeInit}}} & 
            \includegraphics[width=0.18\columnwidth, frame]{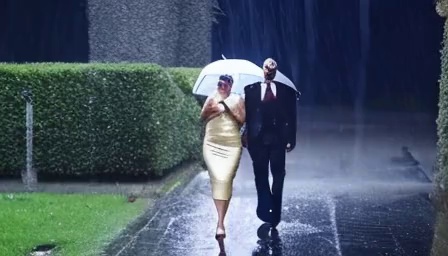} &
            \includegraphics[width=0.18\columnwidth, frame]{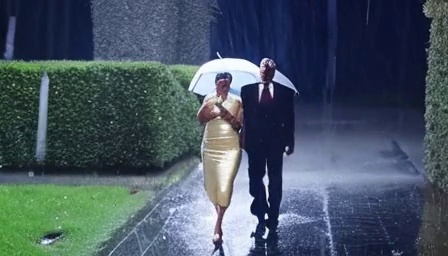} &
            \includegraphics[width=0.18\columnwidth, frame]{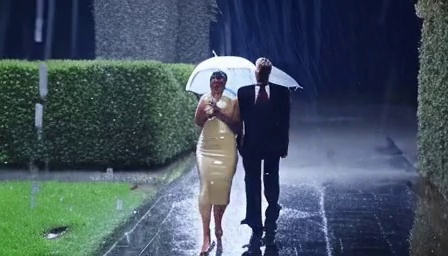} &
            \includegraphics[width=0.18\columnwidth, frame]{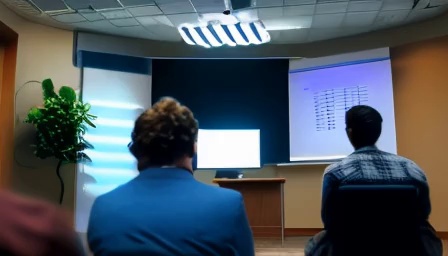} &
            \includegraphics[width=0.18\columnwidth, frame]{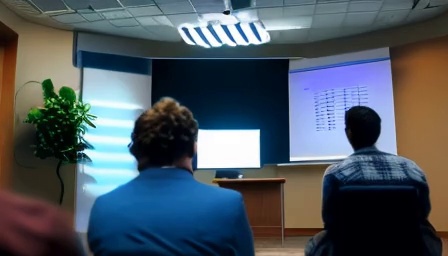} &
            \includegraphics[width=0.18\columnwidth, frame]{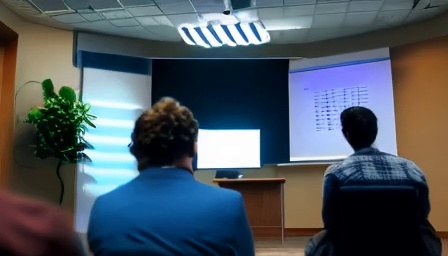}  \\

            \rotatebox{90}{\hspace{2.0mm}{\scalebox{.5}{RagInit}}} & 
            \includegraphics[width=0.18\columnwidth, frame]{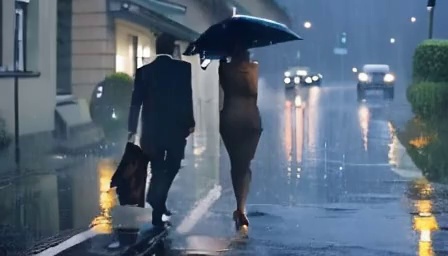} &
            \includegraphics[width=0.18\columnwidth, frame]{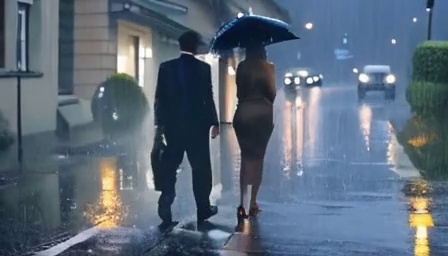} &
            \includegraphics[width=0.18\columnwidth, frame]{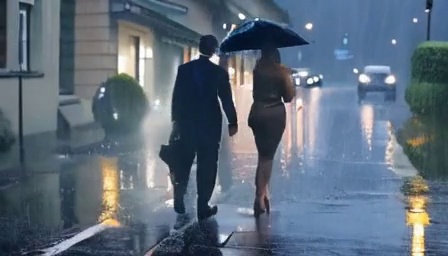} &
            \includegraphics[width=0.18\columnwidth, frame]{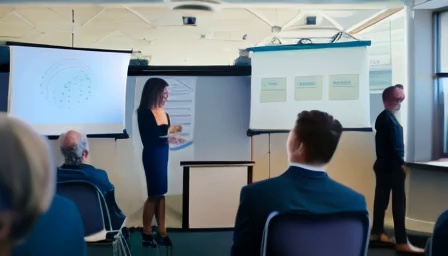} &
            \includegraphics[width=0.18\columnwidth, frame]{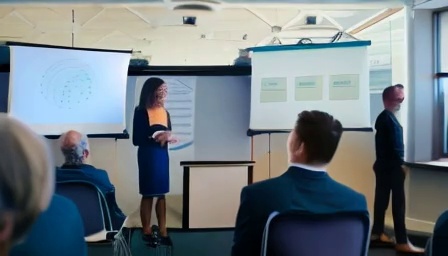} &
            \includegraphics[width=0.18\columnwidth, frame]{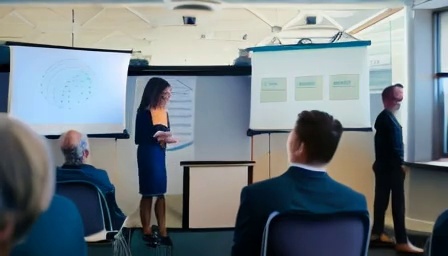}  \\

            \rotatebox{90}{\hspace{2.0mm}{\scalebox{.5}{\textbf{\methodname}}}} & 
            \includegraphics[width=0.18\columnwidth, frame]{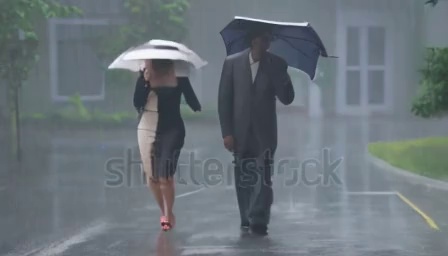} &
            \includegraphics[width=0.18\columnwidth, frame]{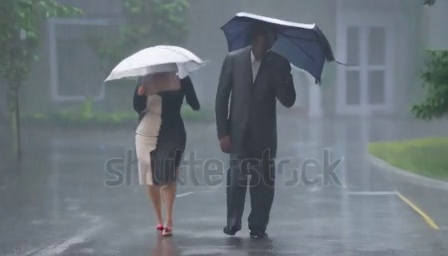} &
            \includegraphics[width=0.18\columnwidth, frame]{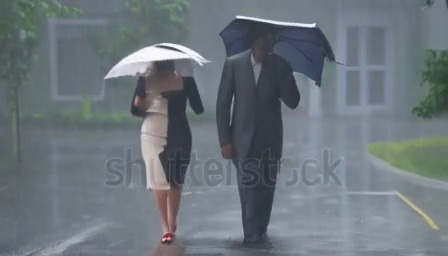} &
            \includegraphics[width=0.18\columnwidth, frame]{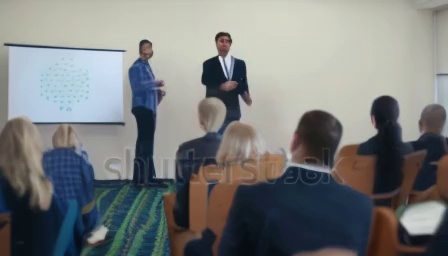} &
            \includegraphics[width=0.18\columnwidth, frame]{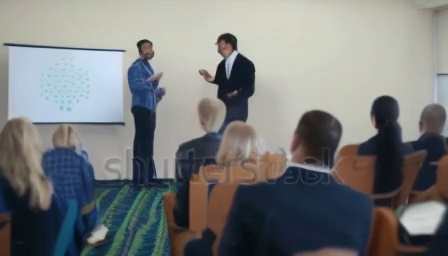} &
            \includegraphics[width=0.18\columnwidth, frame]{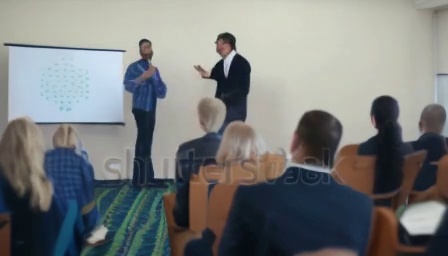} \\
            & \multicolumn{3}{c}{\scalebox{0.5}{\texttt{"A couple [...] going home get caught in a heavy rain."}}} & \multicolumn{3}{c}{\scalebox{0.5}{\texttt{"A person giving a presentation [...]."}}}
    \end{tabular}
    }
    \captionof{figure}{Visual comparison of the different methods. We report the prompt at the bottom.}
    \label{fig:comparison_supp}
\end{table*}

\begin{table*}
    \centering
    \resizebox{\textwidth}{!}{
    \setlength\tabcolsep{0.3pt}
    \footnotesize
    \renewcommand{\arraystretch}{0.3}
    \begin{tabular}{c@{\hskip 1.5pt}ccccc}
            \scalebox{0.5}{Generated} & \multicolumn{5}{c}{\scalebox{0.5}{Retrieved Samples}} \\

            \includegraphics[width=0.18\columnwidth, frame]{figures/comparison/our/000/0000.jpg} &
            \includegraphics[width=0.18\columnwidth, frame]{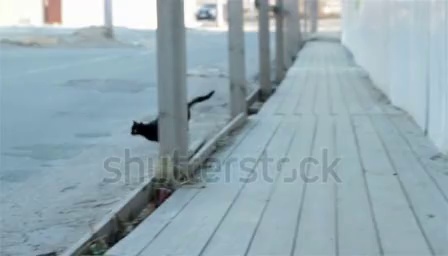} &
            \includegraphics[width=0.18\columnwidth, frame]{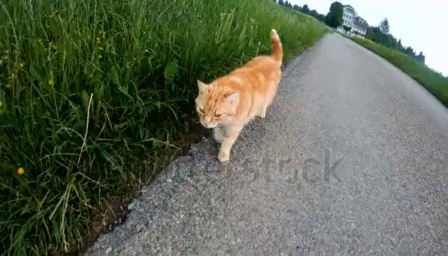} &
            \includegraphics[width=0.18\columnwidth, frame]{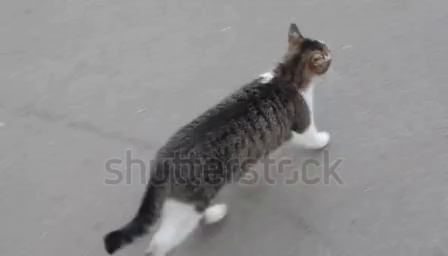} &
            \includegraphics[width=0.18\columnwidth, frame]{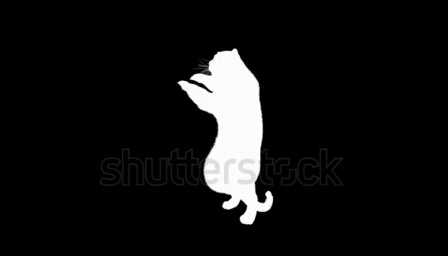} &
            \includegraphics[width=0.18\columnwidth, frame]{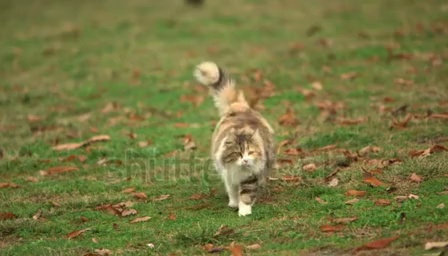}  \\
            \multicolumn{6}{c}{\scalebox{0.5}{\texttt{"A cat running happily."}}} \\

            \includegraphics[width=0.18\columnwidth, frame]{figures/comparison/our/003/0000.jpg} &
            \includegraphics[width=0.18\columnwidth, frame]{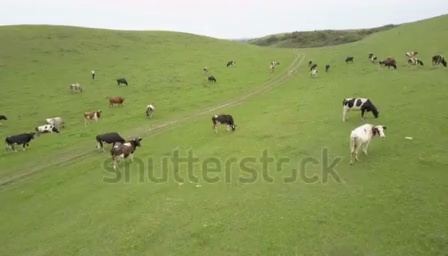} &
            \includegraphics[width=0.18\columnwidth, frame]{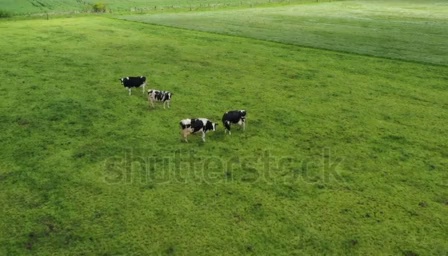} &
            \includegraphics[width=0.18\columnwidth, frame]{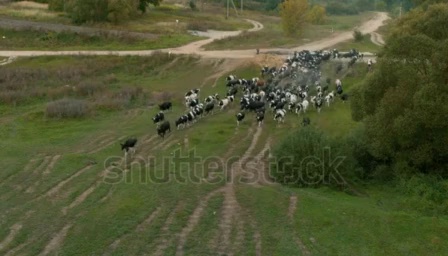} &
            \includegraphics[width=0.18\columnwidth, frame]{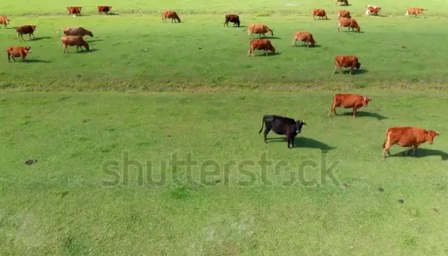} &
            \includegraphics[width=0.18\columnwidth, frame]{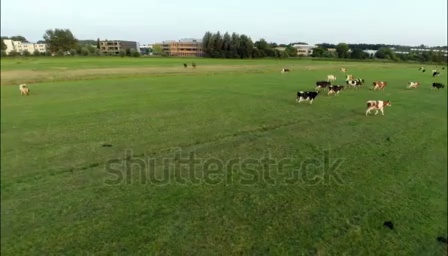}  \\
            \multicolumn{6}{c}{\scalebox{0.5}{\texttt{"A cow running to join a herd of its kind."}}} \\

            \includegraphics[width=0.18\columnwidth, frame]{figures/comparison/our/005/0000.jpg} &
            \includegraphics[width=0.18\columnwidth, frame]{figures/retrieved/retrived/005/0000.jpg} &
            \includegraphics[width=0.18\columnwidth, frame]{figures/retrieved/retrived/005/0001.jpg} &
            \includegraphics[width=0.18\columnwidth, frame]{figures/retrieved/retrived/005/0002.jpg} &
            \includegraphics[width=0.18\columnwidth, frame]{figures/retrieved/retrived/005/0003.jpg} &
            \includegraphics[width=0.18\columnwidth, frame]{figures/retrieved/retrived/005/0004.jpg}  \\
            \multicolumn{6}{c}{\scalebox{0.5}{\texttt{"A zebra running to join a herd of its kind"}}} \\

            \includegraphics[width=0.18\columnwidth, frame]{figures/comparison/our/009/0000.jpg} &
            \includegraphics[width=0.18\columnwidth, frame]{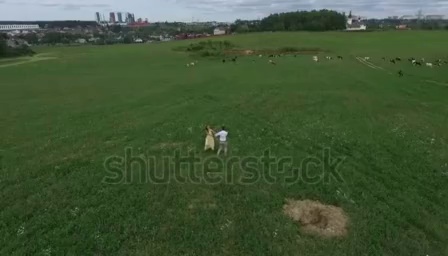} &
            \includegraphics[width=0.18\columnwidth, frame]{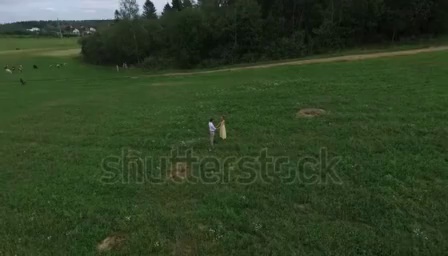} &
            \includegraphics[width=0.18\columnwidth, frame]{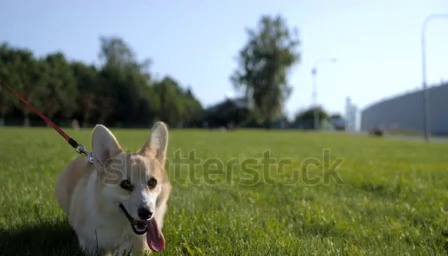} &
            \includegraphics[width=0.18\columnwidth, frame]{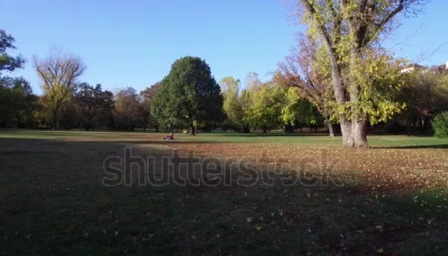} &
            \includegraphics[width=0.18\columnwidth, frame]{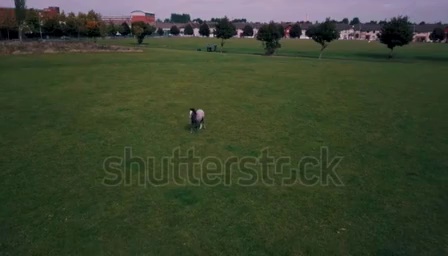}  \\
            \multicolumn{6}{c}{\scalebox{0.5}{\texttt{"A cute happy Corgi playing in park, sunset, zoom out."}}} \\

            \includegraphics[width=0.18\columnwidth, frame]{figures/comparison/our/010/0000.jpg} &
            \includegraphics[width=0.18\columnwidth, frame]{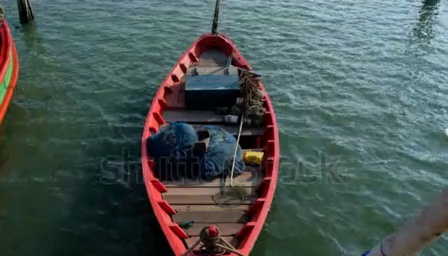} &
            \includegraphics[width=0.18\columnwidth, frame]{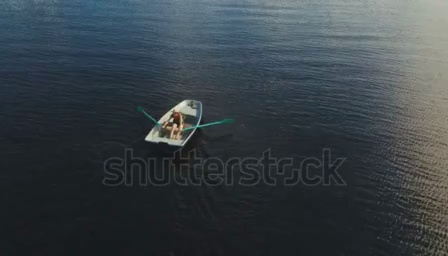} &
            \includegraphics[width=0.18\columnwidth, frame]{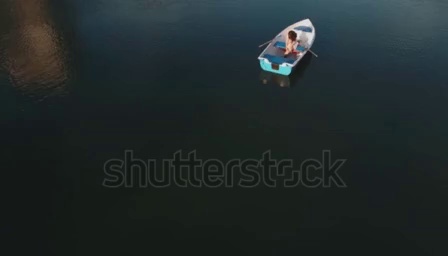} &
            \includegraphics[width=0.18\columnwidth, frame]{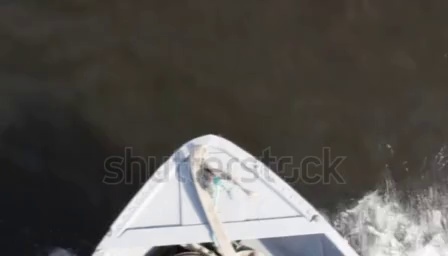} &
            \includegraphics[width=0.18\columnwidth, frame]{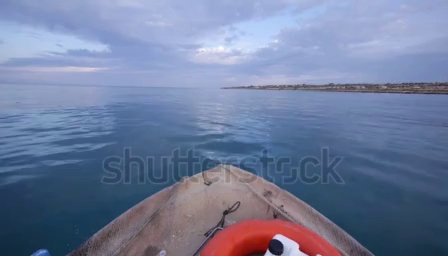}  \\
            \multicolumn{6}{c}{\scalebox{0.5}{\texttt{"A cute raccoon playing guitar in a boat on the ocean."}}} \\

            \includegraphics[width=0.18\columnwidth, frame]{figures/comparison/our/017/0000.jpg} &
            \includegraphics[width=0.18\columnwidth, frame]{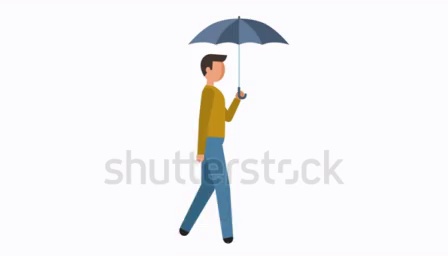} &
            \includegraphics[width=0.18\columnwidth, frame]{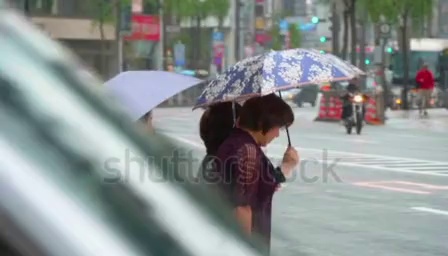} &
            \includegraphics[width=0.18\columnwidth, frame]{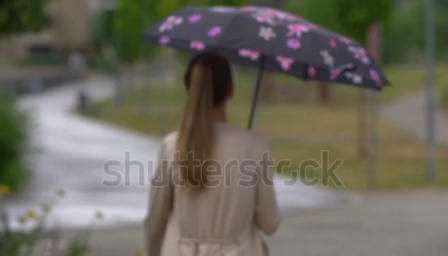} &
            \includegraphics[width=0.18\columnwidth, frame]{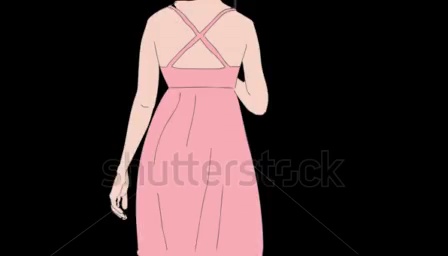} &
            \includegraphics[width=0.18\columnwidth, frame]{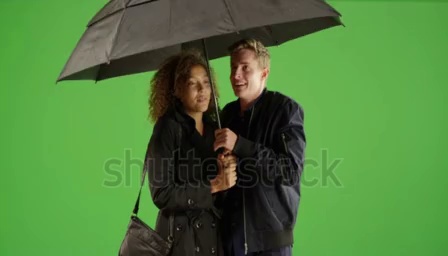}  \\
            \multicolumn{6}{c}{\scalebox{0.5}{\texttt{"A couple in formal evening wear going home get caught in a heavy downpour with umbrellas"}}} \\

            \includegraphics[width=0.18\columnwidth, frame]{figures/comparison/our/023/0000.jpg} &
            \includegraphics[width=0.18\columnwidth, frame]{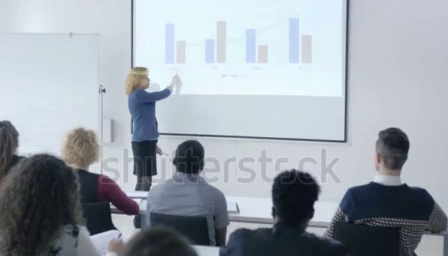} &
            \includegraphics[width=0.18\columnwidth, frame]{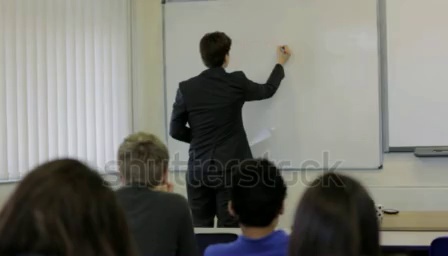} &
            \includegraphics[width=0.18\columnwidth, frame]{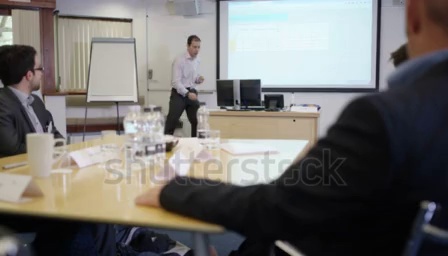} &
            \includegraphics[width=0.18\columnwidth, frame]{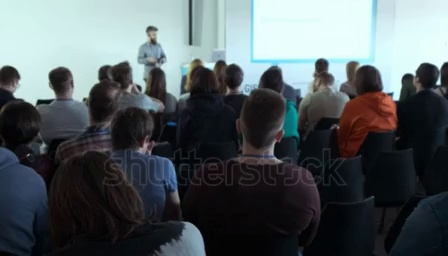} &
            \includegraphics[width=0.18\columnwidth, frame]{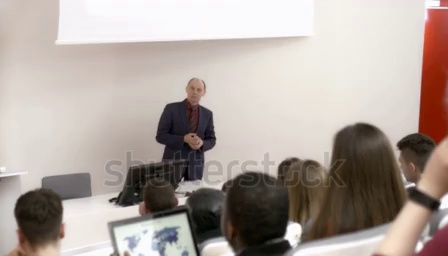}  \\
            \multicolumn{6}{c}{\scalebox{0.5}{\texttt{"A person giving a presentation to a room full of colleagues"}}} \\

            \includegraphics[width=0.18\columnwidth, frame]{figures/comparison/our/029/0000.jpg} &
            \includegraphics[width=0.18\columnwidth, frame]{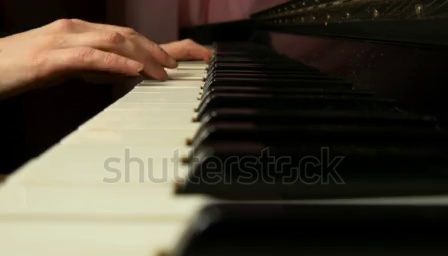} &
            \includegraphics[width=0.18\columnwidth, frame]{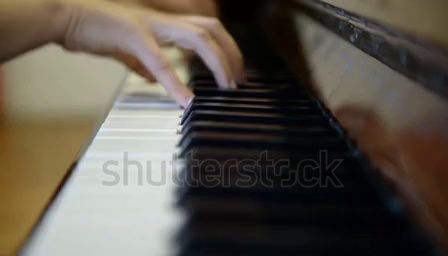} &
            \includegraphics[width=0.18\columnwidth, frame]{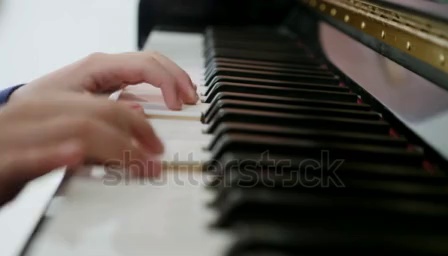} &
            \includegraphics[width=0.18\columnwidth, frame]{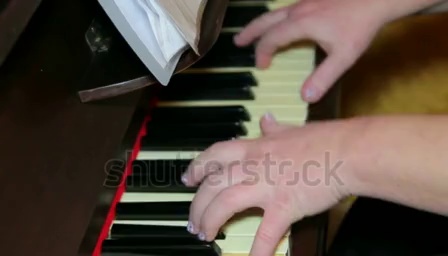} &
            \includegraphics[width=0.18\columnwidth, frame]{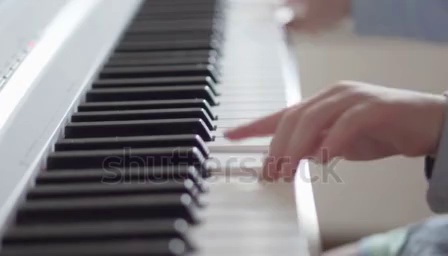}  \\
            \multicolumn{6}{c}{\scalebox{0.5}{\texttt{"A person is playing piano."}}} \\

            \includegraphics[width=0.18\columnwidth, frame]{figures/comparison/our/059/0000.jpg} &
            \includegraphics[width=0.18\columnwidth, frame]{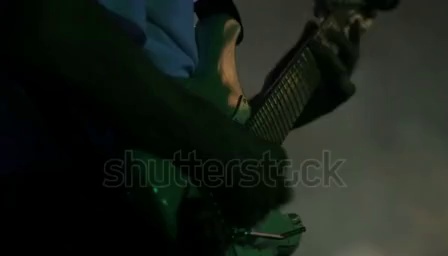} &
            \includegraphics[width=0.18\columnwidth, frame]{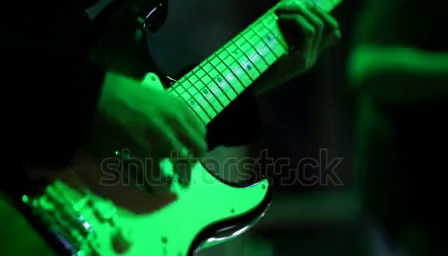} &
            \includegraphics[width=0.18\columnwidth, frame]{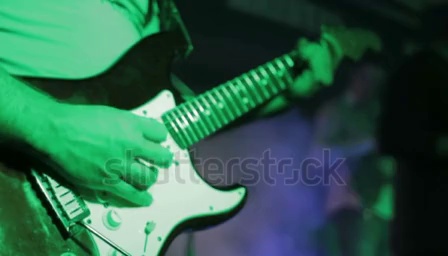} &
            \includegraphics[width=0.18\columnwidth, frame]{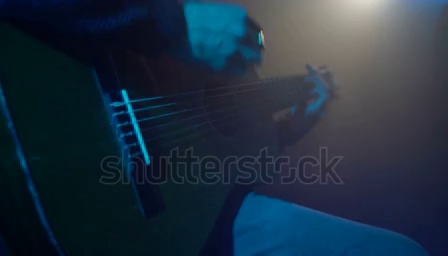} &
            \includegraphics[width=0.18\columnwidth, frame]{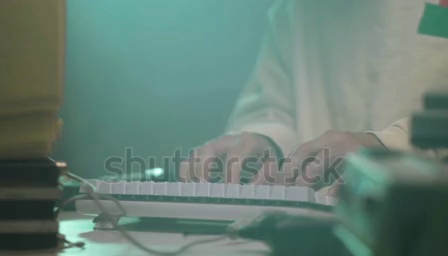}  \\
            \multicolumn{6}{c}{\scalebox{0.5}{\texttt{"Yoda playing guitar on the stage."}}} \\

    \end{tabular}
    }
    \captionof{figure}{We show the first frame of the generated video and the first frame of the 5 retrieved samples used during the generation phase. No clear leakage is present, \ie the model is not simply copy-pasting the output but using it to improve the result.}
    \label{fig:retrived_supp}
\end{table*}

\end{document}